\documentclass[]{fairmeta}

\usepackage{wrapfig}
\usepackage{tabularx}
\usepackage{textcomp}
\usepackage{stfloats}
\usepackage{url}
\usepackage{verbatim}
\usepackage{titlesec}
\usepackage{tocloft}
\usepackage{adjustbox}
\usepackage{multirow}
\usepackage{pifont}
\usepackage{tikz}
\usepackage{comment}
\usepackage{amsmath,amssymb}
\usepackage{colortbl}
\usepackage{color}
\usepackage{booktabs}
\usepackage{hyperref}
\usepackage{graphicx}
\usepackage{subcaption}
\usepackage{multirow}
\usepackage{booktabs}
\usepackage{subcaption}
\definecolor{headerpurple}{HTML}{d8d2fc}
\RequirePackage{xspace}
\makeatletter
\DeclareRobustCommand\onedot{\futurelet\@let@token\@onedot}
\def\@onedot{\ifx\@let@token.\else.\null\fi\xspace}
\usepackage[most]{tcolorbox}
\usepackage{xcolor}
\usepackage{enumitem}

\newcolumntype{C}{>{\centering\arraybackslash}X}

\newcommand{\huggingface}{\raisebox{-1.5pt}{\includegraphics[height=1.05em]{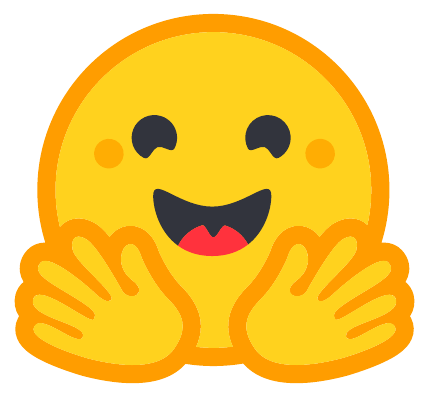}}\xspace}
\newcommand{\github}{\raisebox{-1.5pt}{\includegraphics[height=1.05em]{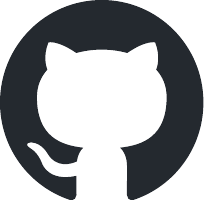}}\xspace}
\newcommand{\homepage}{\raisebox{-1.5pt}{\includegraphics[height=1.05em]{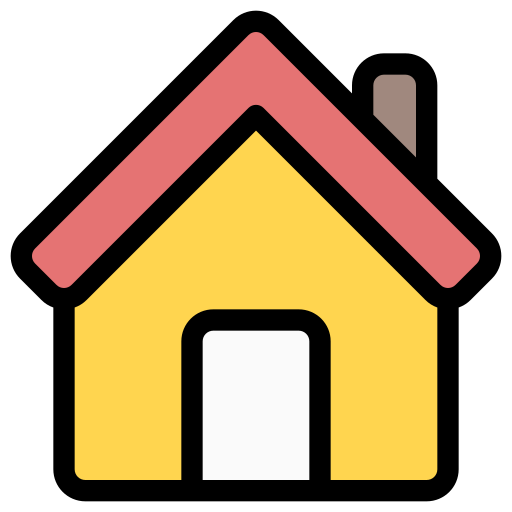}}\xspace}
\newcommand{\modelscope}{\raisebox{-1pt}{\includegraphics[height=0.95em]{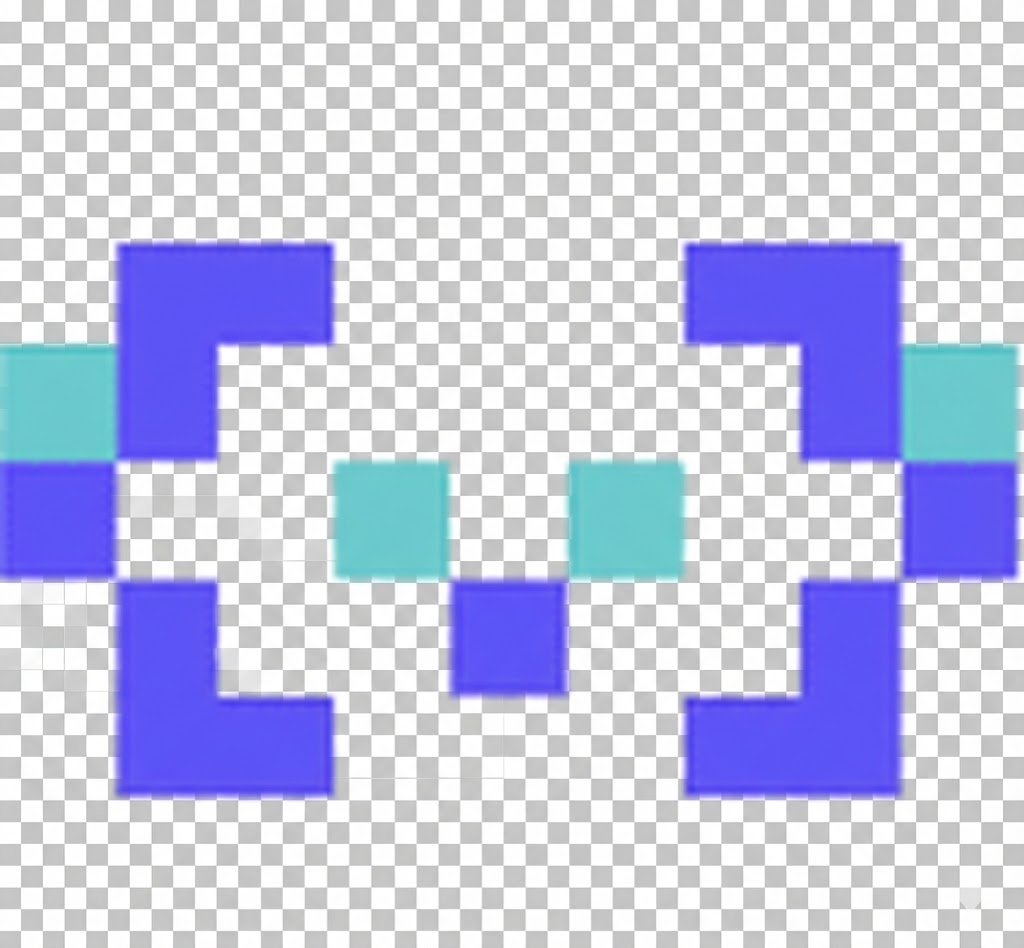}}\xspace}

\makeatother

\definecolor{adptorange}{RGB}{248, 205, 172}
\definecolor{cmpblue}{RGB}{189, 215, 238}
\definecolor{cmpblue}{RGB}{189, 215, 238}

\definecolor{our_red}{RGB}{232,157,160}
\definecolor{our_blue}{RGB}{136,206,230}
\definecolor{our_orange}{RGB}{246,200,168}
\definecolor{our_green}{RGB}{178,211,164}

\definecolor{attn_code0}{RGB}{247,215,200}
\definecolor{attn_code1}{RGB}{238,169,139}
\definecolor{mlp_code0}{RGB}{204,201,221}
\definecolor{mlp_code1}{RGB}{102,95,153}

\definecolor{token_blue}{RGB}{84, 120, 140}

\usepackage{pifont}
\usepackage{bbding}
\usepackage{fontawesome}
\usepackage{xspace}

\usepackage{float}
\newcommand{\our}{{RynnWorld-Teleop}\xspace}
\newcommand{\eg}{\textit{e.g.},\xspace}

\newlength\savewidth

\newcolumntype{x}[1]{>{\centering\arraybackslash}p{#1pt}}
\newcolumntype{y}[1]{>{\raggedright\arraybackslash}p{#1pt}}
\newcolumntype{z}[1]{>{\raggedleft\arraybackslash}p{#1pt}}

\renewcommand{\paragraph}[1]{\vspace{1mm}\noindent\textbf{#1}}
\usepackage{colortbl}
\usepackage{xcolor}
\usepackage{wrapfig}

\renewcommand{\paragraph}[1]{\vspace{1.25mm}\noindent\textbf{#1}}

\usepackage{algorithm}
\usepackage{listings}

\definecolor{codeblue}{rgb}{0.25, 0.5, 0.5}
\definecolor{codekw}{rgb}{0.35, 0.35, 0.75}
\lstdefinestyle{Pytorch}{
    language = Python,
    backgroundcolor = \color{white},
    basicstyle = \fontsize{9pt}{8pt}\selectfont\ttfamily\bfseries,
    columns = fullflexible,
    aboveskip=1pt,
    belowskip=1pt,
    breaklines = true,
    captionpos = b,
    commentstyle = \color{codeblue},
    keywordstyle = \color{codekw},
}

\definecolor{green}{HTML}{009000}
\definecolor{red}{HTML}{ea4335}

\title{\our: An Action-Conditioned World Model for Digital Teleoperation}

\author[* 1, 2, 3]{Haoyu Zhao}
\author[* 1]{Xingyue Zhao}
\author[5]{Hangyu Li}
\author[6]{Biao Gong}
\author[1, 4]{Kehan Li}
\author[\dagger 1, 4]{Siteng Huang}
\author[1, 4]{Xin Li}
\author[\dagger 1]{Deli Zhao}
\author[\dagger 2, 3]{Zhongyu Li}

\affiliation[1]{DAMO Academy, Alibaba Group,}
\affiliation[2]{Hong Kong Embodied AI Lab,}
\affiliation[3]{CUHK,}
\affiliation[4]{Hupan Lab,\\}
\affiliation[5]{Alibaba Group,}
\affiliation[6]{Ant Group}

\contribution[*]{Equal contribution}
\contribution[\dagger]{Corresponding author}

\abstract{
Scaling robot learning requires massive, diverse trajectory data, yet collection is currently bottlenecked by physical teleoperation, where every demonstration binds operator time to specific hardware and workspaces.
We introduce \textbf{digital teleoperation}, a paradigm that decouples data collection from physical constraints by replacing the real robot with a generative world model. In this framework, an operator’s hand-pose stream drives a robot-centric generative world model to synthesize high-fidelity egocentric videos from a single reference image.
The recorded pose stream serves as an embodiment-agnostic action label transferable to any target robot via standard retargeting, yielding complete state-action trajectories for imitation learning independent of physical hardware.
We instantiate this paradigm in \textbf{\our}, a system that integrates depth-aware skeletal conditioning, progressive human-to-robot training on a video Diffusion Transformer (DiT), and streaming autoregressive distillation. This pipeline compresses the generative process into a single-pass inference, enabling 40+ FPS, real-time interactive generation on a single H100 GPU.
Policies trained exclusively on \our-generated data achieve effective zero-shot Sim2Real transfer across dexterous and diverse bimanual tasks. Moreover, augmenting real-world datasets with our digitally teleoperated data consistently improves success rates, demonstrating that \our serves as a high-fidelity, scalable data engine for the next generation of robotic agents.

\begin{center}
    
    \begin{tabular}{ll}
        \homepage  & \url{https://alibaba-damo-academy.github.io/RynnWorld-Teleop.github.io} \\
        \github  & \url{https://github.com/alibaba-damo-academy/RynnWorld-Teleop}\\
        \huggingface & \url{https://huggingface.co/Alibaba-DAMO-Academy/RynnWorld-Teleop} \\
        \modelscope  & \url{https://www.modelscope.cn/models/DAMO_Academy/RynnWorld-Teleop}\\
    \end{tabular}
\end{center}

}

\date{\today}

\begin{document}
\thispagestyle{firstheader}
\maketitle
\pagestyle{empty}

\section{Introduction}
Recent advances in robotics, such as vision-language-action (VLA) models~\citep{black2024pi_0,brohan2022rt,kim2024openvla,li2026causal,zhao2026towards} and world models~\citep{agarwal2025cosmos,ali2025world}, show emerging promise for general-purpose autonomy, yet they remain significantly hindered by data scarcity~\citep{zhao2026high,lepert2025masquerade,zhao2023learning}. Access to such large-scale robot data would enable more straightforward training and potentially unlock a higher performance upper bound. While traditional teleoperation systems~\citep{li2025teleopbench,ze2025twist,zhao2025smap} provide high-quality expert data, they are often confined to fixed laboratory settings and specific objects~\cite{zhao2025physsplat,guo2026articulat3d,zhao2024hfgs}. The immense overhead of manual environment resets and the logistical challenge of procuring diverse real-world objects prevent these systems from capturing the long-tail distribution of interactions. Consequently, achieving the robust generalization required for unstructured environments remains an open challenge that physical platforms alone cannot solve.

\begin{figure}[t]
  \centering
    \includegraphics[width=\linewidth]{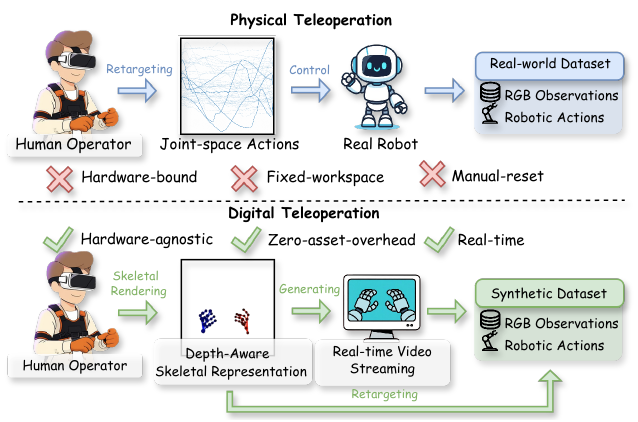}\\
  \caption{\textbf{Physical vs. Digital Teleoperation.}
  (Top) Physical teleoperation binds every demonstration to a real robot and a fixed workspace, capping throughput at operator-hours $\times$ hardware availability. (Bottom) Digital teleoperation replaces the real robot with \our, a real-time action-conditioned world model that synthesizes the egocentric video the robot would have produced from a single reference image, and retargets the same gesture stream into embodiment-specific robot actions. Both pipelines emit synchronized (RGB observations, robotic actions) pairs, so digital teleoperation is drop-in compatible with downstream imitation learning, without ever moving a real robot.
  }
  \label{fig:teaser}
\end{figure}

We propose to decouple operator time from physical infrastructure by replacing the real robot with a generative one. We call this paradigm \textbf{digital teleoperation}: an operator's real-time hand-pose stream is consumed by a generative world model that, conditioned on a single reference image of the target scene, synthesizes the egocentric video that the robot \emph{would have produced} if it had actually executed the gesture (Fig.~\ref{fig:teaser}).
Crucially, because the generated video is grounded in the operator's joint-level hand poses, the recorded gesture itself serves as an embodiment-agnostic, ground-truth action label that can be transferred to any concrete robot embodiment via a standard retargeting pipeline. The generated video then provides the matching visual observations, and the resulting trajectories can be consumed directly by downstream imitation learning, without ever moving a real robot. If digital teleoperation works, robot data collection becomes bound only by operator imagination, not by physical infrastructure.

Two threads of recent work touch on this goal but fail to bridge the gap. Human-to-robot video translation methods~\citep{lepert2025masquerade, li2025h2r, lepert2503phantom, song2025mitty} attempt to ``robotize''  human videos by rendering or translating a robot embodiment into human demonstration frames to bridge the visual gap. While they cross the visual gap, they are inherently passive and observation-only: the underlying robotic action is never produced, and the heavy DiT backbones rule out closed-loop interaction.
Action-conditioned egocentric world models~\citep{hao2026egosim,wang2026hand2world,wang2025precise,xie2026generated,tu2025playerone} go further by synthesizing future frames from human-controlled signals, but they remain human-centric.
The rendered hand is still a human hand, leaving the embodiment gap unbridged.
A genuine digital teleoperation system must simultaneously (i) be robot-centric, so the operator can teleoperate a robot; (ii) be action-grounded, so every generated frame is tied to a recoverable joint-level action signal; and (iii) be real-time, so the operator stays in the control loop and can string complex skills together. No prior framework satisfies all three.

\begin{figure}[t]
  \centering
    \includegraphics[width=0.8\linewidth]{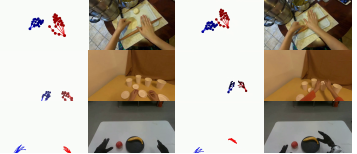}\\
  \caption{\textbf{Depth-Aware Representation.} We bridge the gap between 2D projections and 3D dynamics by rendering hand skeletons with depth-modulated color and size.}
  \label{fig:representation}
\end{figure}

To this end, we present \textbf{\our}, the first system that delivers digital teleoperation in this strict sense, as illustrated in Fig.~\ref{fig:teaser}. \our is a robot-centric, action-conditioned world model that streams high-fidelity interactive video from the egocentric viewpoint at frame rates compatible with real-time control. It is built around three core designs tailored to the requirements above:
\begin{itemize}[nosep, labelsep=0.6em, leftmargin=1.2em,itemindent=0em]
\item \textbf{Depth-aware action representation} (Sec.~\ref{sec:method:rep}, Fig.~\ref{fig:representation}): \our renders 21-joint hand poses with camera-distance-modulated color and radius, supplying explicit 3D cues via a 2D latent conditioning signal.
\item \textbf{Progressive cross-domain training} (Sec.~\ref{sec:method:training}): \our is first pretrained on large-scale egocentric human videos to absorb manipulation priors, and then fine-tuned on paired human–robot data to bridge the embodiment gap.
\item \textbf{Streaming autoregressive distillation} (Sec.~\ref{sec:method:distillation}): We distill a bidirectional teacher into a causal student using a rollout-consistent schedule, enabling stable, real-time, long-horizon generation.
\end{itemize}
\noindent We evaluate \our both as a generative model and, more importantly, as a digital teleoperation system whose output is robot training data. Crucially, policies trained exclusively on \our-generated data achieve effective zero-shot Sim2Real transfer across diverse real-world manipulation tasks, and augmenting real demonstrations with digitally teleoperated data consistently raises success rates. These results demonstrate that \our successfully turns digital teleoperation from an appealing concept into a scalable and efficient tool for the next generation of robot learning.
In summary, our work makes the following contributions:
\begin{itemize}[nosep, labelsep=0.6em, leftmargin=1.2em,itemindent=0em]
\item We formalize \textbf{digital teleoperation} as a new paradigm for robot data generation, where operator hand-pose streams drive a robot-centric world model to generate videos for imitation learning, and identify the three requirements (robot-centric, action-grounded, real-time) that any practical instantiation must satisfy.
\item We present an action-conditioned egocentric world model \textbf{\our} that meets all three requirements, combining a depth-aware skeletal representation, progressive human-to-robot training, and streaming autoregressive distillation into a single 40+ FPS interactive model.
\item We further extend \our into a complete digital teleoperation system that instantiates an arbitrary manipulation scene from a single reference image, generating unbounded action-conditioned rollouts through retargeting, skeletal-conditioned synthesis, and chunked re-anchoring.
\item We show that policies trained purely on data generated by our digital teleoperation system can transfer zero-shot to real robots, and that augmenting real demonstrations with such data further boosts success rates, providing the first empirical evidence that digital teleoperation can serve as a high-fidelity data engine that both substitutes for and amplifies physical teleoperation in robot learning.
\end{itemize}

\section{Related Work}
\subsection{Action-Conditioned Egocentric World Models}
The paradigm of world modeling has recently shifted from latent-space state prediction~\citep{ha2018world,assran2025v,zhao2026rynnworld} to high-fidelity visual simulation, powered by the rapid progress of video diffusion models~\citep{kong2024hunyuanvideo,wang2025wan,yang2024cogvideox}.
Foundation models such as Cosmos~\citep{agarwal2025cosmos} and Genie~\citep{bruce2024genie} demonstrate the potential for physical AI by training on massive video datasets. However, these models typically support only coarse-grained controls, which are insufficient for complex robotic tasks.

To achieve nuanced control, a new class of action-conditioned egocentric world models~\citep{hao2026egosim,wang2026hand2world,xie2026generated,tu2025playerone,goswami2025world,pallotta2025egocontrol,gao2026lome} emerge, focusing on generating first-person videos driven by human action signals.
For example, Hand2World~\citep{wang2026hand2world} distills~\citep{yin2025slow,huang2025self} a bidirectional video diffusion model into a causal autoregressive generator for monocular streaming synthesis.
GeneratedReality~\citep{xie2026generated} conditions future-frame synthesis on hand actions to generate interactive videos. InterDyn~\citep{akkerman2025interdyn} conditions on binary hand masks via a ControlNet~\citep{zhang2023adding}-like branch, while CosHand~\citep{sudhakar2024controlling} generates single-frame hand-object interactions from mask inputs.
Despite these advancements, existing simulators remain predominantly human-centric, primarily focusing on synthesizing human hand movements within virtual scenes.
\our extends this frontier to the robot-centric domain, bridging the gap between large-scale human motion priors and actionable, robot-specific world modeling.

\subsection{Human-to-Robot Video Translation}
Modern video generative models have evolved from text-to-video synthesis~\citep{kong2024hunyuanvideo,wang2025wan,yang2024cogvideox} to complex systems handling multi-modal prompts~\citep{bruce2024genie,jin2024pyramidal}. Within the robotics domain, early efforts~\citep{kling2024,runway2025} focused on visual Human-to-Robot translation.
For example, Phantom~\citep{lepert2503phantom} and Masquerade~\citep{lepert2025masquerade} use inpainting and rendering techniques to overlay robot arms onto estimated human poses. X-Humanoid~\citep{yang2025x} and Mitty~\citep{song2025mitty} leverage Diffusion Transformers for video-to-video translation, while H2R~\citep{li2025h2r} proposes human-to-robot data augmentation for pretraining.
Although crossing the visual domain gap, these approaches are largely passive and observation-only. They perform visual retargeting on pre-recorded videos without providing a mechanism to control future states via actions. This lack of action-conditioning makes them unsuitable for interactive simulators.
\our distinguishes itself by shifting from passive visual translation to active, action-conditioned generation, effectively bridging the gap between human motion priors and actionable robotic data.

\section{\our: An Action-Conditioned Egocentric World Model}
\label{sec:method}

Given a reference image $I_{ref}$ and hand-gesture sequence $\mathcal{P} = \{p_1, \dots, p_T\}$, \our synthesizes high-fidelity robotic egocentric videos $V = \{v_1, \dots, v_T\}$.
To bridge the embodiment gap between human intent and robotic execution, our framework integrates: (i) depth-aware action representation (Sec.~\ref{sec:method:rep}) and the task-specific architecture (Sec.~\ref{sec:method:hands}); (ii) a two-stage progressive training paradigm for human-to-robot knowledge transfer (Sec.~\ref{sec:method:training}); and (iii) an autoregressive distillation process that enables real-time interactive generation (Sec.~\ref{sec:method:distillation}).

\begin{figure}[t]
  \centering
    \includegraphics[width=\linewidth]{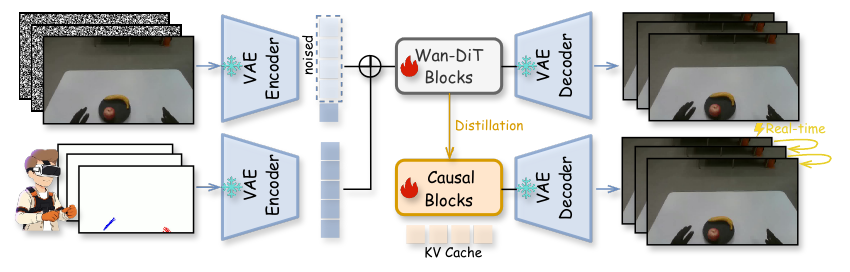}\\
  \caption{\textbf{Overview of \our.} (a) Actions are rendered as depth-aware skeletal videos and encoded into the latent space via a VAE. (b) We expand a pretrained video DiT to incorporate hand-pose conditioning using a distribution-aligned patch embedding branch. (c) The model is distilled into a causal student for interactive, autoregressive generation using a streaming rollout schedule.}
  \label{fig:pipeline}
\end{figure}

\subsection{Preliminaries}
\label{sec:method:preliminaries}
\our builds upon the Wan-I2V architecture~\citep{wang2025wan}, which utilizes a 3D Variational Autoencoder (VAE) and a Transformer-based denoiser $\mathcal{F}_\Theta$. The training objective follows the \textbf{conditional flow matching} (CFM) framework~\citep{lipman2022flow}. Given an initial image latent $z_0 = \mathcal{E}(I_{V})$, the forward process constructs a probability path between data and noise:
\begin{equation}
z_t = (1 - t) z_0 + t \epsilon, \quad \epsilon \sim \mathcal{N}(0, \mathbf{I})
\end{equation}
where $t \in [0, 1]$. The network $v_\Theta$ is trained to predict the velocity field, conditioned on the reference-image latent $z_{ref} = \mathcal{E}(I_{ref})$ and our depth-aware skeletal control latent $c$:
\begin{equation}
\mathcal{L}_{\text{CFM}} = \mathbb{E}_{t, z_0, \epsilon} \left[ \left\| v_\Theta(z_t, t, z_{ref}, c) - (\epsilon - z_0) \right\|_2^2 \right]
\end{equation}
During inference, the video is generated by solving the ODE defined by $v_\Theta$ using a distilled causal student for real-time responsiveness.

\subsection{Depth-Aware Action Representation}
\label{sec:method:rep}
We represent actions as a sequence of skeletal videos derived from 21-joint hand tracking. To resolve the inherent depth ambiguity in standard 2D projections---which is critical for modeling precise hand-object interactions---we employ a \textbf{depth-modulated rendering} technique. Specifically, the depth-encoded color mapping and diameter of each joint and bone are dynamically scaled according to their camera-space depth (as illustrated in Fig.~\ref{fig:representation}). This representation effectively encapsulates articulated hand structures alongside spatial cues essential for manipulation.

To ensure the control signal is compatible with the video generation pipeline, the rendered pose video is projected into the latent space using a pretrained VAE encoder. This yields a control latent $c \in \mathbb{R}^{C \times T \times H \times W}$ that is spatially and temporally aligned with the target video latent, facilitating fine-grained grounding between action and appearance.

\subsection{Action-Conditioned Video Generation}
\label{sec:method:hands}
We build our world model upon a Video Diffusion Transformer (DiT)~\citep{wang2025wan}, conditioning the generation process on hand-pose sequences as a high-dimensional proxy for physical actions.

\noindent \textbf{Pose-to-Video Conditioning.}
Our model follows an image-to-video paradigm: given the reference-image latent $z_{ref}$ and a control latent $c$, the model predicts the subsequent motion. To integrate the pose information, we adopt an additive patch-embedding scheme with \textbf{distribution alignment}. We introduce a dedicated control patch-embedding layer $\operatorname{PatchEmbed}^{c}_{C \rightarrow D}$ in parallel to the original video embedding layer $\operatorname{PatchEmbed}^{z}_{C \rightarrow D}$. Their outputs are fused via a learnable scalar gate $\alpha$:
\begin{equation}
x = \operatorname{PatchEmbed}^{z}_{C \rightarrow D}(z_t) \;+\; \alpha \cdot \operatorname{PatchEmbed}^{c}_{C \rightarrow D}\!\left(\widetilde{c}\right),\quad
\widetilde{c} = \frac{c - \mu_c}{\sigma_c} \cdot \sigma_z + \mu_z,
\label{eq:patchify_addplus}
\end{equation}
where $z_t \in \mathbb{R}^{C \times T \times H \times W}$ is the noisy video latent and $\widetilde{c}$ is the aligned control latent. Since $c$ and $z$ originate from different modalities, they exhibit distinct statistics. We maintain running estimates of the mean and standard deviation $(\mu, \sigma)$ for both signals and align $c$ to the video latent distribution before patchification. This ensures the additive control signal remains statistically compatible with the pretrained video stream throughout training. To preserve the generative prior, $\operatorname{PatchEmbed}^{c}$ is zero-initialized and the gating scalar $\alpha$ is initialized to a small value (\eg $0.1$), allowing the network to gradually incorporate the pose signal without destabilizing the pretrained weights.

\begin{table}[t]
    \centering
    \caption{Statistics of the datasets used for egocentric human pretraining and robotic domain adaptation. We report the total number of frames and the resulting video slices after temporal segmentation.}
    \label{tab:data}
    \small
    \begin{tabular}{lcccc}
        \toprule
        \textbf{Dataset} & \textbf{Type} & \textbf{Seq. Length} & \textbf{Total Frames} & \textbf{Total Slices} \\
        \midrule
        \rowcolor{gray!10} \multicolumn{5}{l}{\textit{Egocentric Human Pretraining}} \\
        VITRA~\citep{li2025scalable} & Human & 25 & 30.7M & 1.23M \\
        EgoDex~\citep{hoque2025egodex} & Human & 81 & 74.0M & 0.91M \\
        \midrule
        \rowcolor{gray!10} \multicolumn{5}{l}{\textit{Robotic Domain Adaptation}} \\
        Ours (Real-Robot) & Robot & 81 & 0.43M & 5.3K \\
        \bottomrule
    \end{tabular}
\end{table}

\subsection{Progressive Cross-Domain Training}
\label{sec:method:training}
To facilitate the transfer of dexterous manipulation skills from human priors to robotic embodiments, we adopt a two-stage training paradigm.
To bridge the human–robot embodiment gap between these two training stages, we convert all data into paired (2D Skeleton, RGB) sequences, and present the per-stage training data statistics in Tab.~\ref{tab:data}.

\noindent
\textbf{Stage 1: Egocentric Human Pretraining.} The model is first trained on large-scale egocentric human video datasets. In this stage, the DiT learns the fundamental dynamics of hand-object interactions and establishes the mapping from skeletal configurations to realistic visual synthesis.

We utilize \textbf{EgoDex}~\citep{hoque2025egodex} and \textbf{VITRA}~\citep{li2025scalable} for pre-training. Specifically, EgoDex is segmented into 81-frame clips. To generate the skeleton guidance, we transform the 3D SE(3) hand poses into the camera frame and project them into 2D space to render joint-link skeleton videos, encompassing both hands and forearms. For VITRA, we extract 25-frame segments and utilize the provided hand action annotations (21 joints in MediaPipe format). These joints are projected into 2D skeleton videos using the camera intrinsics and the camera pose of each clip's initial frame to maintain spatial grounding.

\noindent
\textbf{Stage 2: Robotic Domain Adaptation.} We fine-tune the model on paired teleoperation data, where human gestures are mapped to robotic actions via Inverse Kinematics (IK). This allows the model to function as a robot-centric world model, where human-like intent directly drives robotic execution videos.

To provide the necessary diverse physical interactions for training \our, we collected a large-scale real-world dataset focused on complex bimanual coordination. This dataset consists of four distinct tasks designed to challenge the model's understanding of object dynamics and long-horizon manipulation.

To capture human-centric motion priors effectively, we record the operator's hand movements using a motion capture (MoCap) system, which provides synchronized 2D skeleton sequences. To ensure temporal consistency with the pre-training phase, all robot trajectories are segmented into 81-frame clips. This unified skeleton-based representation, combined with consistent temporal alignment, facilitates an effective transfer of manipulation capabilities from human demonstrations to robotic embodiments.

In total, we collected 1,800 high-quality demonstration episodes. Unlike the task-specific policy training, all 1,800 episodes are utilized to train \our, allowing the generative model to learn a comprehensive world prior from a broad range of physical scenarios. The specific tasks and their respective data distributions are detailed in Table~\ref{tab:teleop_data_stats}.

\begin{table}[t]
\centering
\caption{Statistics of the real-world dataset used for \our training.}
\label{tab:teleop_data_stats}
\begin{tabular}{lp{8cm}c}
\hline
\textbf{Task Name} & \textbf{Description} & \textbf{Number} \\ \hline
\rowcolor{gray!10}Dual Picking & Pick an apple and a banana from a plate to the table. & 500 \\
Block Pushing & Push a block from left to center, and then to the right using the right hand. & 500 \\
\rowcolor{gray!10}Bimanual Lifting & Lift a watermelon pillow with both hands and place it into a tray. & 500 \\
Lid Placement & Place the lid of a cardboard box onto the box. & 300 \\
\rowcolor{headerpurple!60} \textbf{Total} & & \textbf{1,800} \\ \hline
\end{tabular}
\end{table}

\subsection{Autoregressive Distillation}
\label{sec:method:distillation}
To support interactive, closed-loop applications, we distill the bidirectional teacher model into a causal student model~\citep{huang2025self} capable of generating videos autoregressively at the frame level. The process involves reconfiguring the model for streaming inference, followed by a two-stage optimization: a causal flow-matching warm-up and an adversarial distribution matching stage.

\noindent
\textbf{Causal Streaming Architecture.}
The student model employs a causal temporal mask and a fixed-size KV cache (a pre-allocated buffer with in-place writes) to enable low-latency inference. Unlike the teacher, the student's attention at time $t$ is restricted to $\{1, \dots, t\}$. To maintain long-term consistency during autoregressive rollouts, we retain the embedding of $I_{ref}$ as a \textbf{persistent sink token} and append all subsequently generated KV states to the cache. This design preserves the identity and spatial context of the initial reference image throughout generation.

\noindent
\textbf{Causal Flow-Matching Warm-up.}
The training begins with a causal flow-matching~\citep{lipman2022flow} objective to bridge the gap between bidirectional and causal processing. We train the student to predict the velocity field $\mathbf{v}_\theta$ in a frame-causal manner by minimizing the mean squared error:
\begin{equation}
\mathcal{L}_{\text{MSE}} = \mathbb{E}_{t,\boldsymbol{\epsilon}}\left[\left\| \mathbf{v}_\theta(\mathbf{x}_t, t) - (\boldsymbol{\epsilon} - \mathbf{x}_0) \right\|^2\right].
\end{equation}
This phase establishes the student's streaming generation and control-following capabilities using single-step velocity regression, providing a stable initialization for sampling acceleration.

\noindent
\textbf{Distribution Matching Distillation.}
To achieve high-fidelity synthesis in only four denoising steps, we subsequently employ Distribution Matching Distillation (DMD)~\citep{yin2024one}. This refinement utilizes a learned critic and the frozen teacher to provide score-based gradient guidance, pushing the student's output distribution toward that of the teacher. During this phase, the student performs 4-step sampling rollouts. Crucially, we backpropagate gradients through the \textbf{persisted KV cache} across successive chunks. This cross-chunk gradient flow encourages the model to minimize boundary artifacts at transitions, ensuring smooth temporal continuity and enabling the final model to match the visual quality of the teacher in a real-time, interactive loop.

\section{\our as a Digital Teleoperation System}

Sec.~\ref{sec:method} establishes \our as an action-conditioned egocentric world model that turns a hand-pose stream into a high-fidelity, real-time robotic video. To realize the promise of digital teleoperation, however, the world model alone is not enough: it must be embedded in a closed loop that captures operator intent, drives the generator, and produces trajectories directly consumable by a downstream policy. In this section, we describe the surrounding system that turns \our into such a data engine. Sec.~\ref{sec:data_pipeline} details the end-to-end pipeline that converts raw operator motion into paired (video, action) trajectories for imitation learning, while Sec.~\ref{sec:sim_comparison} contrasts this paradigm with traditional physical simulation and clarifies the unique advantages of digital teleoperation as a scalable data source.

\subsection{Data Generation Pipeline for Policy Learning}
\label{sec:data_pipeline}

To facilitate scalable policy training, we develop a systematic pipeline to generate synthetic robot trajectories by leveraging \our as a high-fidelity data engine.

\noindent
\textbf{Retargeting.}
Given the raw 6-DoF poses from the Vive trackers worn on the operator's chest, wrists, and upper arms, we first compute each arm's target end-effector pose via a calibrated coordinate-transform chain that maps the operator's workspace to the robot's workspace through a translational scaling factor. The corresponding joint configuration is then obtained with an iterative damped least-squares (DLS) inverse-kinematics solver, regularized by a null-space shoulder prior derived from the upper-arm trackers to keep the arm in natural, human-like configurations. Joint limits are enforced after each update. This yields, for every operator gesture, a synchronized 54-dimensional robot action vector (dual 7-DoF arms + dual 20-DoF dexterous hands) that is exactly aligned with the hand-pose stream driving \our. The full transform chain, DLS formulation, and null-space objective are detailed in Appendix~\ref{sec:retargeting_appendix}.

\noindent
\textbf{Skeletal-Conditioned Synthesis.} For each source demonstration, we extract the initial RGB frame as the reference image $I_{ref}$. The operator's hand-pose stream is rendered into a 2D depth-aware skeletal sequence $S_{1:T}$ at 16 FPS. By conditioning \our on $I_{ref}$ and $S_{1:T}$, we synthesize the corresponding robotic execution video.
Notably, because $I_{ref}$ is the sole visual specification of the target scene, it can also be user-provided or synthetically edited to instantiate manipulation scenes beyond the training distribution. We validate this generalization capability in Sec.~\ref{sec:ac-wm-eval}.
In either case, the ground-truth robotic actions $a_{1:T}$ are synchronized with the generated video frames. Each action $a_t \in \mathbb{R}^{54}$ is a concatenated vector of the absolute joint positions for the dual 7-DoF robotic arms and dual 20-DoF dexterous hands. These actions are retargeted from the operator's poses during the original teleoperation session, ensuring that synthesized visual observations and robotic states are perfectly aligned for downstream imitation learning.

\noindent
\textbf{Mitigating Drift via Chunked Re-anchoring.} Generating long-horizon videos in a purely autoregressive manner can lead to visual drift and cumulative physical inconsistencies. To ensure the synthetic data remains physically grounded, we adopt a chunk-based generation strategy with a segment length of 81 frames. For the first chunk, the model is initialized with the true starting frame of the demo. For each subsequent chunk, we \textit{re-anchor} the generation by providing the actual egocentric frame from the robot's camera at that specific timestep as the new $I_{ref}$. This periodic grounding ensures that the synthesized environment (e.g., object poses and lighting) remains consistent with the ground-truth action sequence throughout complex, long-horizon tasks.

\subsection{Data Engine Comparison with Traditional Simulation}
\label{sec:sim_comparison}

We compare \our with traditional physical simulators to highlight its advantages as a scalable data engine:

\begin{itemize}[leftmargin=*]
    \item \textbf{Zero 3D Asset Overhead:} Traditional simulation requires explicit 3D meshes and URDFs for every object~\citep{guo2026articulat3d,zhao2025physsplat}. \our instantiates the environment implicitly from a single reference image $I_{ref}$, completely bypassing the manual asset modeling bottleneck.
    \item \textbf{No Visual Domain Gap:} Simulated rendering inherently suffers from a reality gap~\citep{zhao2026towards}. \our synthesizes frames directly within the real-world pixel distribution, naturally capturing complex lighting, reflections, and sensor noise without domain randomization.
    \item \textbf{Implicit Physics vs. Manual SysID:} Simulators require tedious System Identification (friction, mass) and struggle with deformable objects. By pre-training on large-scale human videos, \our absorbs physical ``common sense,'' generating plausible contact dynamics without explicit differential equations.
\end{itemize}

\section{Experiments}
\subsection{Training Details}

We adopt Wan2.2-TI2V-5B~\citep{wang2025wan} as our base model. To align the base model with the visual characteristics of our target environments (\eg lighting, specific objects), we first perform standard text-and-image-to-video (TI2V) fine-tuning as a warm-up. This stage does not involve action conditioning. We train the full model for 2,000 steps with a learning rate of $2 \times 10^{-5}$ on 64 NVIDIA GPUs, allowing the model to adapt to the specific visual domain of the human and robotic datasets.

\noindent
\textbf{Action-Conditioned Training.}
We introduce the depth-aware skeletal conditioning branch and explore two training paradigms to adapt the model: Low-Rank Adaptation (LoRA) and full-parameter Supervised Fine-Tuning (SFT). Both paradigms follow a two-stage progressive schedule:
\begin{itemize}[leftmargin=*]
    \item \textbf{Stage 1 (Egocentric Human Pretraining):} We initialize the model with the pretrained Wan-2.2-TI2V-5B backbone and learn fundamental manipulation dynamics on large-scale human datasets.
    \begin{itemize}[leftmargin=*]
        \item \textit{LoRA-based:} We optimize the patch embedding layers and LoRA weights (rank 64) with a learning rate of $2 \times 10^{-5}$.
        \item \textit{Full-parameter SFT:} We perform full-parameter fine-tuning with a learning rate of $2 \times 10^{-5}$, employing a 200-step warmup and EMA (decay $0.999$) to stabilize the learning of interaction priors.
    \end{itemize}

    \item \textbf{Stage 2 (Robotic Domain Adaptation):} We fine-tune the Stage 1 checkpoints on 1,800 paired human-robot episodes to bridge the embodiment gap.
    \begin{itemize}[leftmargin=*]
        \item \textit{LoRA-based:} We continue training the LoRA weights and control branch with a learning rate of $1 \times 10^{-5}$.
        \item \textit{Full-parameter SFT:} We refine the full model parameters with a learning rate of $1 \times 10^{-5}$, incorporating EMA (decay $0.999$) and a 200-step warm-up to ensure stability across complex bimanual coordination tasks.
    \end{itemize}
\end{itemize}

\textbf{Autoregressive Distillation.} Finally, the bidirectional teacher is converted into a causal student for real-time inference:
\begin{itemize}[leftmargin=*]
    \item \textbf{Causal Flow-Matching Warm-up:} The student model undergoes a warm-up phase to establish temporal causality with a learning rate of $1 \times 10^{-5}$.
    \item \textbf{Adversarial Distillation (DMD):} For few-step interactive generation, we apply DMD with a generative learning rate of $2 \times 10^{-6}$ and a critic learning rate of $5 \times 10^{-7}$.
\end{itemize}

\subsection{Digital Teleoperation System Evaluation}

\noindent
\textbf{Experimental Setup.}
Experiments are conducted on a TIANJI M6 mobile robot equipped with dual arms and dual WUJI dexterous hands. An egocentric RealSense D435i captures the visual stream. Please refer to Appendix~\ref{sec:body} for hardware and the inverse kinematics solver details.

As shown in Fig.~\ref{fig:tasks}, to demonstrate generalization across diverse manipulation challenges, we evaluate our method on four distinct tasks requiring varying levels of bimanual coordination:

\begin{figure}
  \centering
    \includegraphics[width=\linewidth]{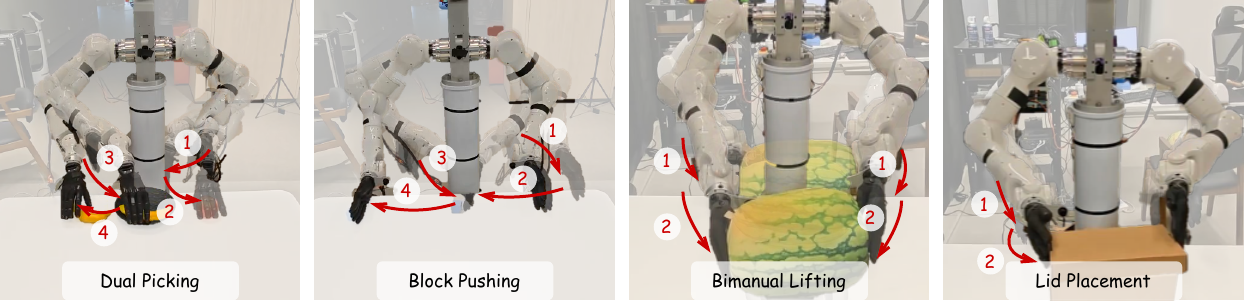}\\
  \caption{\textbf{Task illustration}. We design four manipulation tasks for real-world evaluation.}
  \label{fig:tasks}
\end{figure}

\begin{itemize}[labelsep=0.6em, leftmargin=1.2em,itemindent=0em]
    \item \textbf{(1) \textit{Dual Picking}}: The robot uses its left arm to pick an apple and its right arm to pick a banana from a plate, sequentially placing both objects onto the tabletop.

    \item \textbf{(2) \textit{Block Pushing}}: A sequential pushing task where the left arm pushes a large block from the left zone to the center, after which the right arm takes over to push it to the designated right target zone.

    \item \textbf{(3) \textit{Bimanual Lifting}}: A heavy-load coordination task where both arms must synchronously lift a watermelon plush from the table and accurately place it into a tray.

    \item \textbf{(4) \textit{Lid Placement}}: A precision-oriented task requiring the robot to pick up a lid and accurately align it to cover a cardboard box.
\end{itemize}

\begin{table*}[t]
 \centering
 \scriptsize
 \caption{\textbf{Real-world policy performance.} Success rates (\%) over 35 consecutive trials per task. \textbf{Color-Aug} is a simple photometric-augmentation control that expands the 300 real episodes with brightness, contrast, saturation, and hue jitter under the same 300-episode budget as our generated data. For H2R~\citep{li2025h2r}, we pair the generated robotized observations with the same retargeted action labels used in our pipeline.}
 \label{tab:sim2real}
 \setlength{\tabcolsep}{2pt}

 \begin{tabularx}{\textwidth}{l l CCCCC}
 \toprule
 \textbf{Method} & \textbf{Data Source} & \textbf{Dual Picking} & \textbf{Block Pushing} & \textbf{Bimanual Lifting} & \textbf{Lid Placement} & \textbf{Avg.} \\
 \midrule
 DP~\citep{chi2025diffusion} & 0 Real + 300 \our & 57.14 & 60.00 & 60.00 & 40.00 & 54.29 \\
 \rowcolor{gray!10} DP & 300 Real & 82.86 & 85.71 & 88.57 & 57.14 & 78.57 \\
 DP & 300 Real + 300 Color-Aug & 85.71 & \textbf{88.57} & 91.43 & \textbf{68.57} & 83.57 \\
 \rowcolor{headerpurple!60} DP & 300 Real + 300 \our & \textbf{88.57} & \textbf{88.57} & \textbf{94.29} & 65.71 & \textbf{84.29} \\
 \midrule
 $\pi_{0.5}$~\citep{intelligence2025pi_} & 0 Real + 300 \our & 74.29 & 77.14 & 71.43 & 34.29 & 64.29 \\
 \rowcolor{gray!10} $\pi_{0.5}$ & 300 Real & 94.29 & \textbf{100.00} & 94.29 & 42.86 & 82.86 \\
 $\pi_{0.5}$ & 300 Real + 300 Color-Aug & \textbf{97.14} & 97.14 & 94.29 & 60.00 & 87.14 \\
 \rowcolor{headerpurple!60} $\pi_{0.5}$ & 300 Real + 300 \our & \textbf{97.14} & 97.14 & \textbf{100.00} & \textbf{62.86} & \textbf{89.29} \\
 \midrule
 $\pi_0$~\citep{black2024pi_0} & 0 Real + 300 \our & 68.57 & 82.86 & 77.14 & 28.57 & 64.29 \\
 \rowcolor{gray!10} $\pi_0$ & 300 Real & 88.57 & 94.29 & 91.43 & 34.29 & 77.14 \\
 $\pi_0$ & 300 Real + 300 Color-Aug & 91.43 & 97.14 & 97.14 & 42.86 & 82.14 \\
 \rowcolor{gray!10} $\pi_0$ & 300 Real + 300 H2R~\citep{li2025h2r} & 91.43 & 94.29 & 94.29 & 45.71 & 81.43 \\
 $\pi_0$ & 300 Real + 300 \our (w/o Human Pre-training) & 78.57 & 87.14 & 87.14 & 31.43 & 71.07 \\
 \rowcolor{headerpurple!60} $\pi_0$ & 300 Real + 300 \our & 94.29 & \textbf{100.00} & 97.14 & 54.29 & 86.43 \\
 \rowcolor{headerpurple!60} $\pi_0$ & 300 Real + 600 \our & \textbf{97.14} & \textbf{100.00} & \textbf{100.00} & \textbf{74.29} & \textbf{92.86} \\
 \bottomrule
 \end{tabularx}
\end{table*}

These tasks collectively challenge the model's proficiency in dual-arm synergy, temporal sequencing, and long-horizon interaction.
For each task, we evaluate the model's generalization by applying significant randomization to the initial environment state. This includes varying the 6-DoF poses of task-relevant objects (\eg fruits, containers) in terms of both workspace coordinates and axial rotations.
The primary evaluation metric is the \textbf{Success Rate}, defined as the percentage of successful completions over 35 consecutive real-world trials per task. A trial is considered successful if the robot reaches the target state within 120 seconds. At $n=35$ the $95\%$ Wilson interval for a single cell has a half-width of about $\pm 16$ points near a $50\%$ rate, narrowing to about $\pm 5$ points at ceiling, so we base our conclusions on the four-task average and on effects that replicate across policies rather than on individual cells.

\noindent
\textbf{Policy Learning via Generative Data Scaling.}
Beyond visual fidelity, \our functions as a generative data engine for digital teleoperation: it scales robot training sets by translating unconstrained human gestures into photorealistic execution trajectories that are perfectly synchronized with ground-truth actions.

We evaluate this capability by augmenting three state-of-the-art policies: Diffusion Policy (DP)~\citep{chi2025diffusion}, $\pi_{0.5}$~\citep{intelligence2025pi_}, and $\pi_0$~\citep{black2024pi_0}, and include a matched-budget \textbf{Color-Aug} control that expands the same 300 real episodes with photometric jitter. As shown in Tab.~\ref{tab:sim2real}, augmenting 300 real episodes with 300 \our-generated episodes improves the four-task average for all three policies (DP $78.57\!\rightarrow\!84.29$, $\pi_{0.5}$ $82.86\!\rightarrow\!89.29$, $\pi_0$ $77.14\!\rightarrow\!86.43$). The gains are largest on \textit{Lid Placement}, the most precision-critical task, where $\pi_{0.5}$ and $\pi_0$ both improve by $+20$ points over real-only ($42.86\!\rightarrow\!62.86$ and $34.29\!\rightarrow\!54.29$).

Color-Aug alone raises the average by $+5.00$ (DP), $+4.28$ ($\pi_{0.5}$), and $+5.00$ ($\pi_0$), showing that part of the gain is generic data expansion. \our further improves over this matched-budget control by $+0.72$, $+2.15$, and $+4.29$, respectively, with the largest residual gain again on \textit{Lid Placement} (\eg $\pi_0$: $42.86\!\rightarrow\!54.29$). A sign test over non-tied task--policy cells supports this trend: excluding tied cells, \our outperforms the matched-budget Color-Aug control in 7 of 8 task--policy cells, indicating that the benefit is most evident where precise alignment is required. The H2R~\citep{li2025h2r} comparison under $\pi_0$, paired with the same retargeted actions, yields a comparable generic-augmentation gain ($77.14\!\rightarrow\!81.43$, $+4.29$) but remains below \our ($+9.29$). In contrast, replacing \our with the no-human-pretraining generator reduces the $\pi_0$ average to $71.07$, below real-only, consistent with its generative collapse (Tab.~\ref{tab:exp}). Thus, the benefit of \our is not generic to any generated video; it depends on robot-centric pose-conditioned generation and human egocentric pretraining.

In the \textbf{generated-data-only} condition, where policies are trained only on 300 generated episodes, all three policies attain non-trivial success rates---$54.29$ for DP and $64.29$ for both $\pi_{0.5}$ and $\pi_0$---but remain below their real-data counterparts, with average gaps of $-24.3$ (DP), $-18.6$ ($\pi_{0.5}$), and $-12.9$ ($\pi_0$). Generated data therefore does not yet replace physical teleoperation, but is grounded enough to support non-trivial control and serves as a useful augmentation source. Notably, $\pi_0$ trained \textbf{solely on 300 \our-generated episodes} still reaches \textbf{82.86\%} in \textit{Block Pushing} and 77.14\% in \textit{Bimanual Lifting}, and scaling to \textbf{300 Real + 600 \our} lifts its four-task average to \textbf{92.86}. By transforming low-cost human gesture sequences into high-value robotic training data, \our provides a scalable solution to the data scarcity bottleneck in real-world robotics.

\begin{figure}[t]
  \centering
    \includegraphics[width=\linewidth]{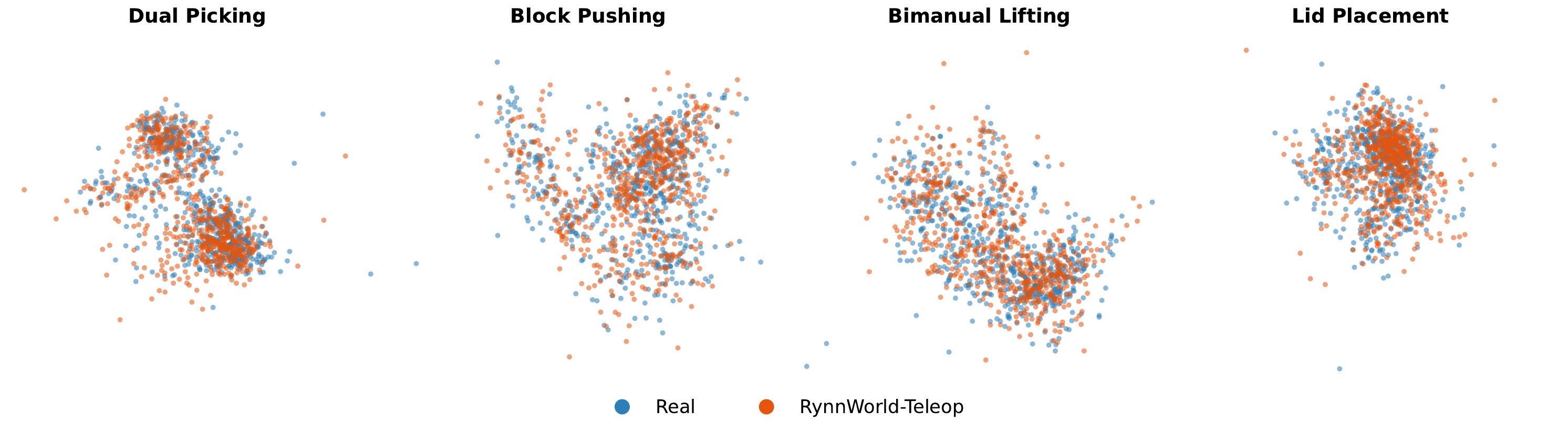}
  \caption{\textbf{Feature Distribution Analysis via t-SNE}. We visualize the high-dimensional feature embeddings of real-world trajectories and \our-generated trajectories.}
  \label{fig:tsne}
\end{figure}

\noindent
\textbf{Feature Distribution Analysis.}
To further investigate the visual and semantic alignment between real-world robot trajectories and those synthesized by \our, we perform a feature distribution analysis using t-SNE. We randomly sample 1000 frames each from our real-world demonstration set and the \our-generated dataset. Features are extracted using a pre-trained I3D~\citep{carreira2017quo} to capture high-level semantic information.

As shown in Fig.~\ref{fig:tsne}, the distribution of \our-generated data significantly overlaps with the real-world data. This alignment indicates that our generative world model captures the coarse semantic statistics of robotic manipulation, including complex hand-object interactions and environmental lighting, and is consistent with the transfer we observe in the policy learning experiments. We note that distributional overlap in an I3D feature space is necessary but not sufficient for physical correctness: it can be satisfied by frames that are semantically plausible yet inconsistent with the paired action, so we treat it as supporting evidence rather than as a validation of observation--action consistency.

\noindent
\textbf{Latency Analysis for Real-time Control.}
To validate the real-time capability of \our (Causal), we evaluate its end-to-end inference latency on a single NVIDIA H100 GPU. Our distilled causal student model, optimized with a 4-step flow matching schedule, achieves a high throughput of \textbf{40.0~fps} at $480 \times 832$ resolution.

The average per-frame latency is approximately $25$\,ms, which can be decomposed into three primary stages:
\begin{itemize}[leftmargin=*]
    \item \textbf{Skeletal Action Encoding ($\sim$5\%):} Pre-processing and VAE encoding of the depth-aware skeletal sequence to provide the conditioning latent.
    \item \textbf{Causal DiT Denoising ($\sim$72\%):} Autoregressive latent generation via 4-step DiT inference, leveraging a sliding-window KV cache for temporal efficiency.
    \item \textbf{Visual Decoding ($\sim$23\%):} VAE decoding from the latent space to the final RGB pixel space.
\end{itemize}

This performance results in an effective interactive frequency of $\sim$40\,Hz, which significantly exceeds the typical $2$--$10$\,Hz frame rates of existing action-conditioned world models \cite{wang2026hand2world, akkerman2025interdyn}. Crucially, this throughput matches or exceeds the standard $30$\,Hz frequency of real-world robotic cameras, thereby minimizing the sensing-to-actuation gap and ensuring a fluid, responsive experience during digital teleoperation.

\begin{table*}[t]
 \small
 \centering
\caption{\textbf{Quantitative results of \our.} We report visual quality metrics and inference speeds. \textbf{Text-} and \textbf{Action-conditioned World Models}, as well as the \textbf{Ablation Study}, are evaluated on \textbf{EgoDex-Test}, while \textbf{Robot-Specific} evaluations are conducted on \textbf{Robotic-Test}.}
    \label{tab:exp}
    \setlength{\tabcolsep}{12pt}
    \begin{tabular}{lccccc}
    \toprule
    \textbf{Method} & \textbf{PSNR} $\uparrow$ & \textbf{SSIM} $\uparrow$ & \textbf{LPIPS} $\downarrow$ & \textbf{FVD} $\downarrow$ & \textbf{FPS} $\uparrow$  \\
    \midrule
    \multicolumn{6}{c}{\textbf{Text-conditioned World Models}} \\
    \rowcolor{gray!10} CogVideoX-1.5-I2V-5B~\citep{yang2024cogvideox}  & 18.22 & 0.786 & 0.322 & 2790 &  0.8 \\
    Wan-2.2-TI2V-5B~\citep{wang2025wan}  & 18.61 & 0.772 & 0.373 & 1998 & 2.8 \\
    \rowcolor{gray!10} Wan-2.1-I2V-14B~\citep{wang2025wan}  & 18.08 & 0.735 & 0.418  & 1540 & 0.3   \\
    Wan-2.2-I2V-14B~\citep{wang2025wan}  & 21.05 & 0.816 & 0.265  & 1337 & 0.3  \\
    \rowcolor{gray!10}Wan-2.2-TI2V-5B (SFT)~\citep{wang2025wan}  & 20.93 & 0.806 & 0.282 & 1223 & 2.8   \\

    \midrule[\heavyrulewidth]
    \multicolumn{6}{c}{\textbf{Action-conditioned World Models}} \\
    \rowcolor{gray!10} InterDyn~\citep{akkerman2025interdyn} & 21.47 & 0.831 & 0.279 & 655 & 2.9  \\
    CosHand~\citep{sudhakar2024controlling} & 18.14 & 0.785 & 0.406 & 1527 & 0.8 \\
    \rowcolor{gray!10}Mask2IV~\citep{li2026mask2iv} & 21.50 & 0.836 & 0.219 & 1650 & 0.9  \\
    \rowcolor{headerpurple!60} \textbf{\our} (LoRA~\citep{hu2022lora})  & 26.08 & 0.876 & 0.151 & 585 & 2.8 \\
    \rowcolor{headerpurple!60} \textbf{\our} (SFT)  & \textbf{26.78} & \textbf{0.887} & \textbf{0.119} & \textbf{550} & 2.8 \\
    \rowcolor{headerpurple!60} \textbf{\our-Causal}  & 22.25 & 0.830 & 0.207 & 1226 & \textbf{40.0} \\

    \midrule[\heavyrulewidth]
    \multicolumn{6}{c}{\textbf{Ablation Study (LoRA)}} \\
    \rowcolor{gray!10} Concatenation Fusion & 19.69 & 0.821 & 0.260 & 1191 & 2.8  \\
    w/o Depth Modulation (plain 2D skeleton) & 19.03 & 0.846 & 0.191 & 830 & 2.8 \\
    \rowcolor{gray!10} w/o Human Pre-training & 17.81 & 0.763 & 0.453 & 2598 &  2.8 \\
    w/o DMD (Causal) & 19.25 & 0.777 & 0.244 & 1338 & \textbf{40.0}  \\
    \rowcolor{gray!10} w/o Causal Warm-up (Causal) & 14.26 & 0.688 & 0.408 & 2150 & \textbf{40.0}  \\

    \midrule[\heavyrulewidth]
    \addlinespace[-1.1pt]
    \midrule[\heavyrulewidth]
    \multicolumn{6}{c}{\textbf{Action-conditioned Robot-Specific World Models}} \\
    \rowcolor{headerpurple!60} \textbf{\our}  & 22.53 & 0.898 & 0.148  & 763  & 2.8 \\
    \rowcolor{headerpurple!60} \textbf{\our-Causal}  & 18.66 & 0.743 & 0.249 & 1534 & \textbf{40.0}  \\

    \bottomrule
    \end{tabular}
    \vspace{-1mm}
\end{table*}

\subsection{Action-Conditioned World Model Evaluation}
\label{sec:ac-wm-eval}

\noindent
\textbf{Experimental Setup.}
To provide a comprehensive assessment of \our's generative capabilities, we establish two distinct evaluation benchmarks as reported in Tab.~\ref{tab:exp}:

\begin{itemize}[labelsep=0.6em, leftmargin=1.2em,itemindent=0em]
    \item \textbf{EgoDex-Test (Human-Centric Domain)}: For evaluating general interaction priors and conducting the ablation study, we utilize the EgoDex dataset~\citep{hoque2025egodex}. We randomly sample 50 video sequences from the official EgoDex test set, each consisting of 81 frames at 16 FPS. These sequences cover a wide variety of unseen human-object interactions. Using this representative subset ensures an unbiased evaluation of how well the model transfers large-scale human manipulation priors to action-conditioned synthesis.

    \item \textbf{Robotic-Test (Robot-Specific Domain)}: To assess the model's performance on actual robotic embodiments and real-world laboratory environments, we hold out a specific \textbf{Robotic-Test} set. This set consists of 20 video sequences (81 frames each) randomly selected from our self-collected teleoperation data, ensuring they were not seen during the Stage 2 Robotic Adaptation training. These sequences encompass all four task categories (Dual Picking, Block Pushing, Bimanual Lifting, and Lid Placement) to verify that \our can maintain high visual fidelity and temporal consistency across diverse robotic manipulation scenarios.
\end{itemize}

To evaluate action-conditioned video generation, we employ a comprehensive set of metrics: \textbf{PSNR} and \textbf{SSIM} for structural fidelity, \textbf{LPIPS} for perceptual similarity, and \textbf{FVD} to assess temporal coherence.
For FVD, we use the standard I3D backbone to evaluate the distribution alignment and temporal coherence of the 81-frame rollouts. All metrics are reported as the mean over the test sequences of the respective benchmark (50 for EgoDex-Test, 20 for Robotic-Test).
We also report \textbf{FPS} measured on a single NVIDIA H100 GPU.

\noindent
\textbf{On statistical reporting.} We report means and avoid per-cell significance tests, and the sample sizes are small enough that this deserves to be stated plainly. Generative metrics are averaged over 20--50 sequences, so differences of a few tenths of a PSNR point should not be over-interpreted; we draw conclusions only from the large margins (\eg the FVD gap between \our and the vanilla SFT baseline, or the collapse of the w/o Human Pre-training variant). Real-robot success rates rest on 35 trials per task, so the $95\%$ Wilson binomial interval for a single cell has a half-width that depends on where the rate sits: about $\pm 16$ points near $50\%$, about $\pm 12$ points at $83\%$, about $\pm 8$ points at $94\%$, and about $\pm 5$ points at ceiling. A single cell is therefore weak evidence on its own, and we never pool trial counts across policies, since different architectures cannot be treated as draws from a common success distribution. Accordingly, in Tab.~\ref{tab:sim2real} we base our conclusions on the four-task average and on effects that replicate across policies. Our only significance statement is a sign test over the twelve task--policy cells comparing \our with the matched-budget Color-Aug control: seven wins, one loss, and four ties, which over the eight non-tied cells yields a one-sided $p=0.035$ (two-sided $p=0.070$). We treat this as supporting rather than definitive evidence, and we report all twelve cells so that the pattern can be judged directly.

\begin{figure}[t]
  \centering
    \includegraphics[width=\linewidth]{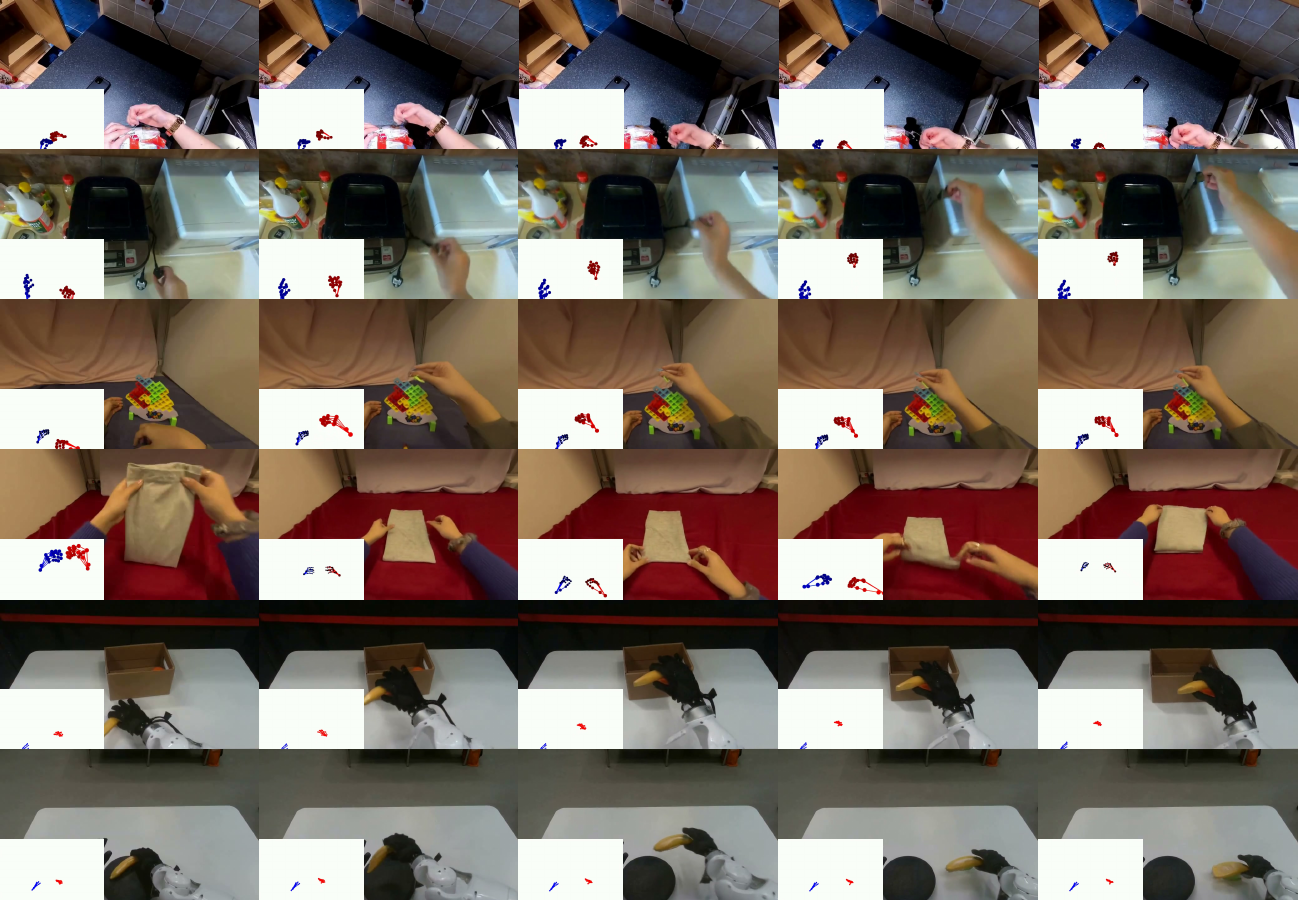}\\
    \caption{\textbf{Qualitative Results of \our.} Our model effectively maps a single hand-gesture stream to both human and robotic embodiments. By pre-training on large-scale egocentric human videos, \our absorbs rich manipulation priors and successfully transfers this knowledge to robotic execution. The synthesized videos maintain high fidelity and temporal coherence.}
  \label{fig:vis}
\end{figure}

\begin{figure}%
  \centering
    \includegraphics[width=\linewidth]{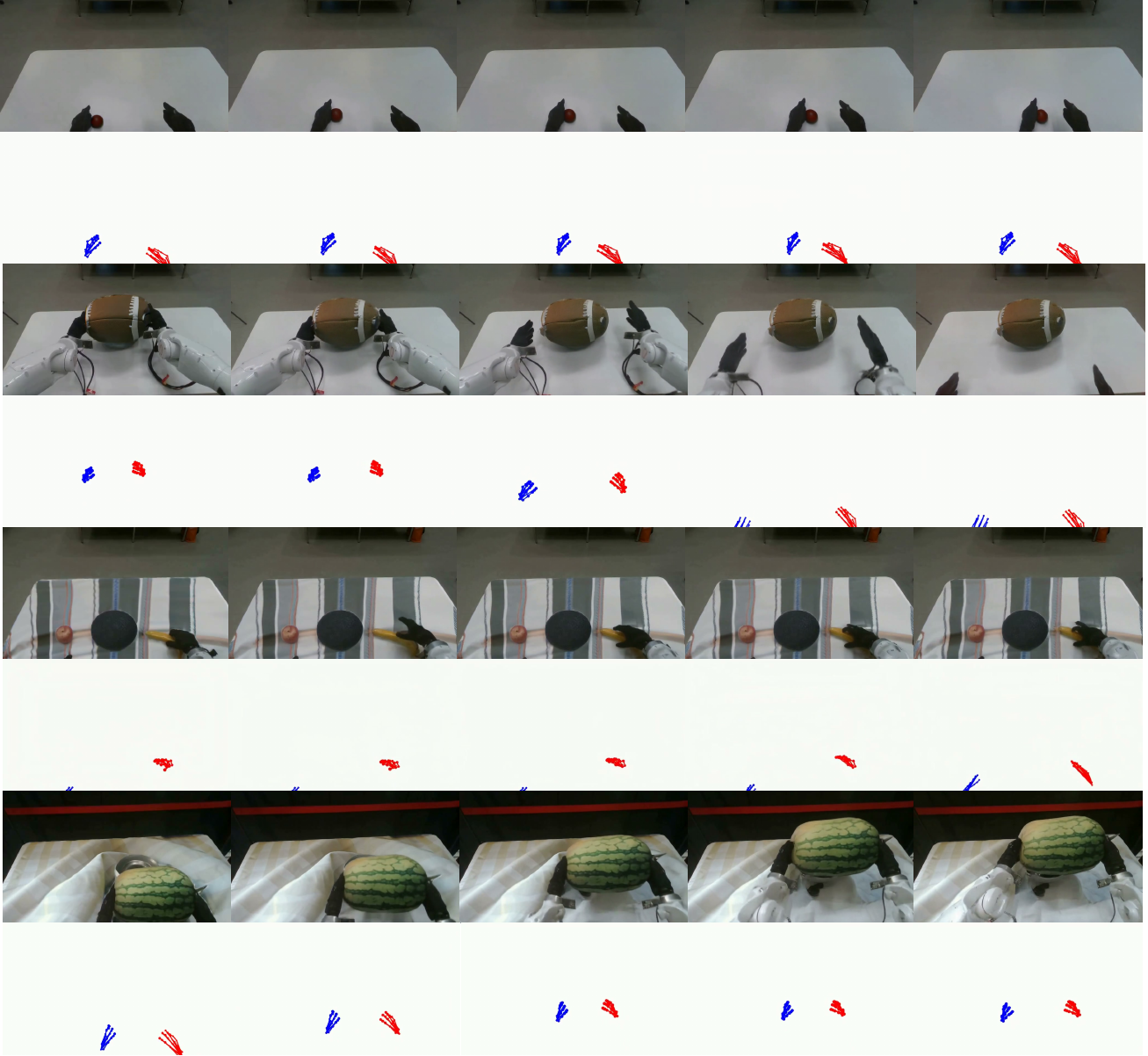}\\
  \caption{\textbf{Generalization to Out-of-Distribution (OOD) Visual States}.
  \our generalizes along two axes that are absent from training: (i) unseen objects (top two rows), where the manipulated item in the reference frame is replaced by a novel category or shape;
  and (ii) unseen backgrounds (bottom two rows), where the tabletop texture is swapped for an environment not seen during training. In all cases, the modified reference image is obtained via off-the-shelf image editing, without any real-world object or scene modification. \our preserves high visual fidelity, temporal coherence, and physically plausible dynamics under both object-level and background-level shifts.
  }
  \label{fig:ood_vis}
\end{figure}

\noindent
\textbf{Action-Conditioned World Modeling.}
As shown in Tab.~\ref{tab:exp}, our method significantly outperforms both general-purpose I2V models and action conditioned video generation models.

General I2V baselines like Wan~\citep{wang2025wan} and CogVideoX~\citep{yang2024cogvideox}, while possessing strong texture priors, lack the mechanism to respond to fine-grained hand gestures, leading to poor temporal coherence in manipulation (FVD $>1300$).
\textbf{Notably, a vanilla SFT baseline that directly fine-tunes Wan-2.2-TI2V-5B on the same robotic and human datasets (denoted as SFT in Tab.~\ref{tab:exp}) still trails \our by a wide margin (FVD 1223 vs. 585; PSNR 20.93 vs. 26.08).} This comparison underscores that simple supervised fine-tuning is insufficient for mastering complex robotic dexterity. In contrast, \our's superior performance stems from our specialized depth-aware pose representation and distribution-aligned conditioning, which provide more explicit and effective action grounding.

Compared to human-centric models like CosHand~\citep{sudhakar2024controlling} and Mask2IV~\citep{li2026mask2iv}, \our effectively bridges the embodiment gap by accurately rendering the complex robotic embodiments. Our distilled Causal student model maintains this high fidelity while supporting interactive generation, facilitating responsive closed-loop applications, supported by the visualizations in Fig.~\ref{fig:vis}.

\begin{figure}[t]
  \centering
  \begin{minipage}{0.49\linewidth}
    \centering
    \vspace{-10pt}
    \includegraphics[width=\linewidth]{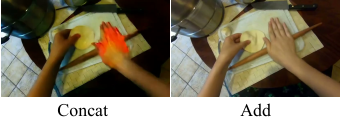}
    \caption{\textbf{Conditioning Strategy.} Compared to concatenation, our distribution-aligned additive scheme (Eq.~\ref{eq:patchify_addplus}) preserves pre-trained weights, ensuring stable synthesis.}
    \label{fig:ablation_fusion}
  \end{minipage}
  \hfill
  \begin{minipage}{0.49\linewidth}
    \centering
    \includegraphics[width=\linewidth]{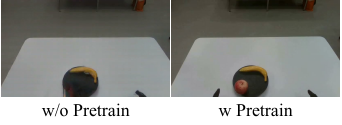}
    \caption{\textbf{Effect of Pretraining.} Stage 1 pretraining provides a robust interaction prior. Without it, the model fails to maintain visual quality when fine-tuned on limited robotic data.}
    \label{fig:ablation_pretrain}
  \end{minipage}
\end{figure}

\noindent
\textbf{Generalization to Out-of-Distribution (OOD) Visual States.}
A key advantage of digital teleoperation is that a new manipulation scene can be instantiated from a single reference image. To validate this claim, beyond the closed-set benchmarks above, we assess \our's ability to generalize to scenes that lie outside the training distribution. Specifically, we consider two complementary axes of visual OOD generalization: unseen objects and unseen backgrounds. For unseen objects, we replace the manipulated item in the reference image with a category or shape that never appears in the training data (\eg substituting a wooden block with a red sphere, or a watermelon plush with a rugby ball). For unseen backgrounds, we swap the tabletop texture in the reference image (\eg overlaying a patterned tablecloth on a plain white table). In both settings, the modified reference image is produced via off-the-shelf image editing, without any real-world object or scene modification. As shown in Fig.~\ref{fig:ood_vis}, \our maintains temporal consistency and faithfully executes the action-conditioned trajectory under both object-level and background-level shifts. These results suggest that \our learns an interaction prior that is largely decoupled from specific object textures and background appearances, which is a prerequisite for instantiating target scenes from a single reference image. We are careful to scope what this establishes: the evaluation is at the level of generation fidelity, and we do not yet train a policy on data generated from edited reference images and deploy it against the corresponding novel physical objects. That policy-level OOD experiment is the decisive test of whether digital teleoperation can collect data for objects that are not physically available, and we identify it as the most valuable direction for future work.

\subsection{Ablation Study}
All ablations use the LoRA variant on EgoDex-Test, so each row of the ablation block in Tab.~\ref{tab:exp} should be read against the \our (LoRA) reference (PSNR 26.08, FVD 585) rather than the Robot-Specific block. Qualitative comparisons for the fusion and pretraining ablations are provided in Fig.~\ref{fig:ablation_fusion} and Fig.~\ref{fig:ablation_pretrain}.

\noindent
\textbf{Depth-modulated rendering.}
Rendering joints and bones with a fixed color and radius, while preserving the same 21-joint topology and 2D projection, removes only the camera-space depth cue and sharply degrades pixel fidelity and temporal coherence relative to the \our (LoRA) reference (PSNR $26.08 \rightarrow 19.03$, FVD $585 \rightarrow 830$, LPIPS $0.151 \rightarrow 0.191$), while SSIM drops comparatively little ($0.876 \rightarrow 0.846$)---consistent with depth errors corrupting fine-grained along-axis motion rather than global structure. This confirms that our advantage over 2D-conditioned baselines is not attributable solely to the pretrained backbone or the human-pretraining stage. The cue matters most when the gesture is near-degenerate under projection, \eg approaching an object along the camera axis, which a plain 2D skeleton cannot distinguish from retreating.

\noindent
\textbf{Conditioning Strategy: Addition vs. Concatenation.}
We investigate the impact of different fusion strategies for integrating pose control. We compare our distribution-aligned additive scheme (Eq.~\ref{eq:patchify_addplus}) with the standard \textbf{Concatenation Fusion}, where control latents are appended to the noisy video latents. As shown in Tab.~\ref{tab:exp}, Concat Fusion results in a significant performance drop, with FVD increasing from 585 to 1191. This indicates that direct concatenation disrupts the pre-trained latent distribution of the base DiT, leading to unstable synthesis. Qualitatively, as shown in Fig.~\ref{fig:ablation_fusion}, our additive approach with zero-initialized gating effectively preserves the generative priors, ensuring better visual fidelity while maintaining precise control.

\noindent
\textbf{Value of Human Egocentric Pre-training.}
To validate our two-stage training paradigm, we evaluate a variant \textbf{w/o Human Pre-training}, which is directly fine-tuned on robotic data. Quantitative results in Tab.~\ref{tab:exp} show a severe performance collapse (FVD 2598, LPIPS 0.453), as the limited robotic data is insufficient to teach the model complex hand-object physics from scratch.
As visualized in Fig.~\ref{fig:ablation_pretrain}, without the interaction priors from human videos, the model fails to synthesize realistic hand structures and exhibits severe ``ghosting'' and blurred textures, ultimately leading to a loss of object permanence where the robot effector and the target apple vanish entirely during manipulation.
This confirms that transferring knowledge from large-scale human egocentric datasets is crucial for bridging the embodiment gap and achieving high-quality robotic world modeling.

\noindent
\textbf{Effectiveness of Sequential Distillation.}
We evaluate the necessity of our two-stage distillation curriculum in Tab.~\ref{tab:exp}. Comparing the final model with the Flow-Matching-only baseline (\textbf{w/o DMD}) reveals a significant quality gap (PSNR 22.25 vs. 19.25). While flow-matching establishes the basic causal architecture, it lacks the high-fidelity refinement of the teacher. More importantly, we observe that directly applying DMD~\citep{yin2024one} to the bidirectional teacher without the causal warm-up phase (\textbf{w/o Causal Warm-up}) leads to severe training instability and blurred textures. This confirms that the first stage (Causal Warm-up) is essential for bridging the structural gap between bidirectional and causal processing, providing a stable initialization that allows the subsequent DMD~\citep{yin2024one} phase to successfully inherit the teacher’s complex hand-object interaction priors in just 4 steps.

\section{Conclusion}
In this paper, we presented \our, a generative digital teleoperation framework that bridges the gap between unconstrained human gestures and precise robotic execution videos. By introducing a depth-aware hand pose representation and a progressive two-stage training paradigm, our model successfully transfers interaction priors from large-scale human datasets to specific robotic embodiments. To support real-time interactive use cases, we further distilled the bidirectional teacher model into a causal student via a novel streaming autoregressive distillation procedure, ensuring spatiotemporal consistency over extended horizons. Experimental results demonstrate that \our not only synthesizes high-fidelity, action-consistent interaction videos but also serves as a powerful generative data engine. Augmenting real-world robotic datasets with trajectories synthesized by \our significantly improves the success rate and generalization of manipulation policies. We believe this work provides a scalable path toward training general-purpose robotic agents by leveraging human intuition as a digital control signal, and we hope digital teleoperation will become a standard ingredient in scalable robot data pipelines.

\noindent
\textbf{Limitation.}
While \our successfully demonstrates digital teleoperation as a viable data engine, several limitations remain.
First, while the depth-modulated rendering captures 3D spatial dynamics, the model occasionally struggles with complex physical phenomena such as fine-grained liquid dynamics or the manipulation of highly deformable objects.
Addressing these cases will likely require richer training data covering such interactions.
Second, bridging the embodiment gap currently requires per-platform fine-tuning, which limits the paradigm's scalability across robot fleets. We view cross-embodiment foundation world models, possibly conditioned on robot kinematic descriptors, as a promising direction.

\newpage
\bibliographystyle{assets/plainnat}
\bibliography{paper}

\newpage
\beginappendix

\begin{figure}
  \centering
    \includegraphics[width=\linewidth]{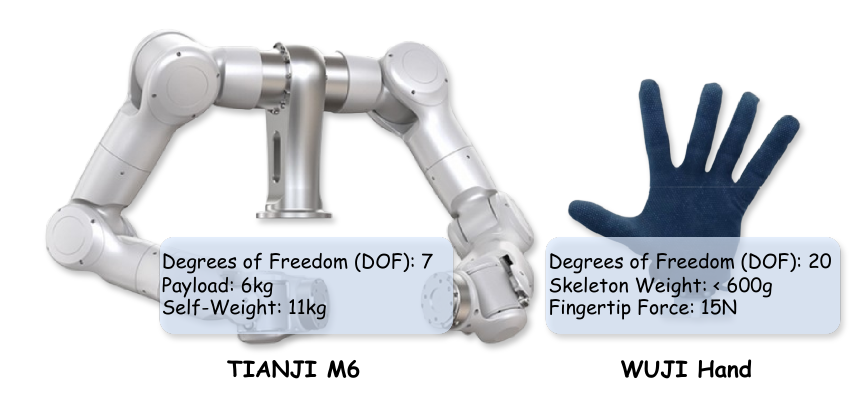}\\
  \caption{\textbf{Hardware Infrastructure.} The evaluation utilizes our standard robotic setup: a bimanual TIANJI M6 system equipped with WUJI hands. Detailed specifications including degrees of freedom (DoF), payload, and fingertip force are summarized to ensure the reproducibility of our manipulation benchmark.}
  \label{fig:body}
\end{figure}

\section{Real Robot System Setup}
\label{sec:body}
Our real robot is built on the TIANJI M6 and WUJI Hand, as shown in Fig.~\ref{fig:body}.
The policy’s inference frequency is set at 50 Hz. The commands are sent with a delay kept between 18 and 30 milliseconds. The low-level interface operates at a frequency of 500 Hz, ensuring smooth real-time control. The communication between the control policy and the low-level interface is realized through LCM (Lightweight Communications and Marshalling).

We collect real-world demonstration data through a teleoperation system.
Our hardware setup consists of dual TIANJI 7-DoF robotic arms and dual WUJI 20-DoF dexterous hands, yielding a total of 54 degrees of freedom.
For arm control, the human operator wears five HTC Vive trackers (mounted on the chest, both wrists, and both upper arms). The system computes wrist-to-chest relative transforms at 100-120 Hz and feeds them into a Pinocchio-based inverse kinematics solver running in a separate process. The resulting joint commands are further smoothed by a Ruckig trajectory generator with velocity, acceleration, and jerk constraints before being sent to the robot arms at 200 Hz.

\begin{figure}
  \centering
    \includegraphics[width=0.4\linewidth]{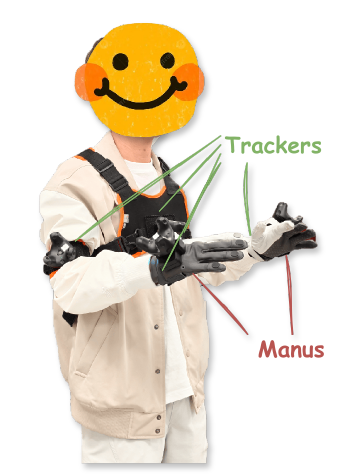}\\
  \caption{\textbf{Operator setup} for real-world data collection.}
  \label{fig:operator}
\end{figure}

For hand control, the operator, as shown in Fig.~\ref{fig:operator}, wears Manus data gloves. The raw glove signals are converted to a 21-point MediaPipe hand skeleton format and retargeted to the 20-DoF WUJI hand joint space via a dedicated retargeting module, with an exponential moving average filter applied for motion smoothing.

\section{Retargeting in \our}
\label{sec:retargeting_appendix}
We provide here the full retargeting pipeline summarized in Sec.~\ref{sec:data_pipeline}.
Given the raw poses from the Vive trackers, we first compute the target end-effector pose for each arm via a coordinate transform chain:
\begin{equation}                                                                       \mathbf{T}_{\text{target}} = \mathbf{T}_{\text{base}} \cdot \text{Scale}(\mathbf{T}_{\text{chest}}^{-1} \cdot \mathbf{T}_{\text{wrist}}) \cdot \mathbf{T}_{\text{ee}},
\end{equation}
where $\mathbf{T}_{\text{chest}}^{-1} \cdot \mathbf{T}_{\text{wrist}}$ extracts the wrist-to-chest relative pose in tracker space, the $\text{Scale}(\cdot)$ operator scales only the translation component by a factor $s$ (\eg $s=1.5$) to map the operator's workspace to the robot's workspace, and $\mathbf{T}_{\text{base}}$, $\mathbf{T}_{\text{ee}}$ are fixed calibration transforms that align the tracker coordinate frame with the robot base frame and end-effector frame, respectively.

We solve the inverse kinematics using an iterative damped least-squares (DLS) method. At each iteration, the 6D task-space error $\mathbf{e} \in \mathbb{R}^6$ (position and orientation) is computed, and the joint update is:
\begin{equation}                                                                       \Delta \mathbf{q} = \mathbf{J}^{\#}_{\lambda} \, \mathbf{e},                           \end{equation}                                                                         where $\mathbf{J}^{\#}_{\lambda}$ is the damped pseudo-inverse obtained via SVD:       \begin{equation}                                                                       \mathbf{J} = \mathbf{U} \boldsymbol{\Sigma} \mathbf{V}^\top, \quad                     \mathbf{J}^{\#}_{\lambda} = \mathbf{V} \, \text{diag}\!\left(\frac{\sigma_i}{\sigma_i^2 + \lambda^2}\right) \mathbf{U}^\top,                                       \end{equation}                                                                         with an adaptive damping factor $\lambda = \lambda_{\min} + \frac{0.01}{1 + \sigma_{\max}}$ that increases near singularities.

To produce natural arm configurations, we exploit a \textbf{null-space shoulder prior} derived from the upper-arm trackers. Specifically, the arm tracker position is transformed into the robot frame to yield a reference elbow position, from which  we solve a partial IK (least-squares on a $3 \times 2$ sub-Jacobian of joints $q_0, q_1$) to obtain shoulder reference values $\mathbf{q}_{\text{shoulder}}^{\text{ref}}$. These are injected as a weighted null-space task:
\begin{equation}
\Delta \mathbf{q} \leftarrow \Delta \mathbf{q} + (\mathbf{I} - \mathbf{J}^{\#}_{\lambda} \mathbf{J}) \, w \, (\mathbf{q}^{\text{ref}} - \mathbf{q}),
\end{equation}
where $w=0.5$ for the shoulder joints. Joint limits are enforced by hard clipping after each integration step.

\begin{figure}
  \centering
    \includegraphics[width=\linewidth]{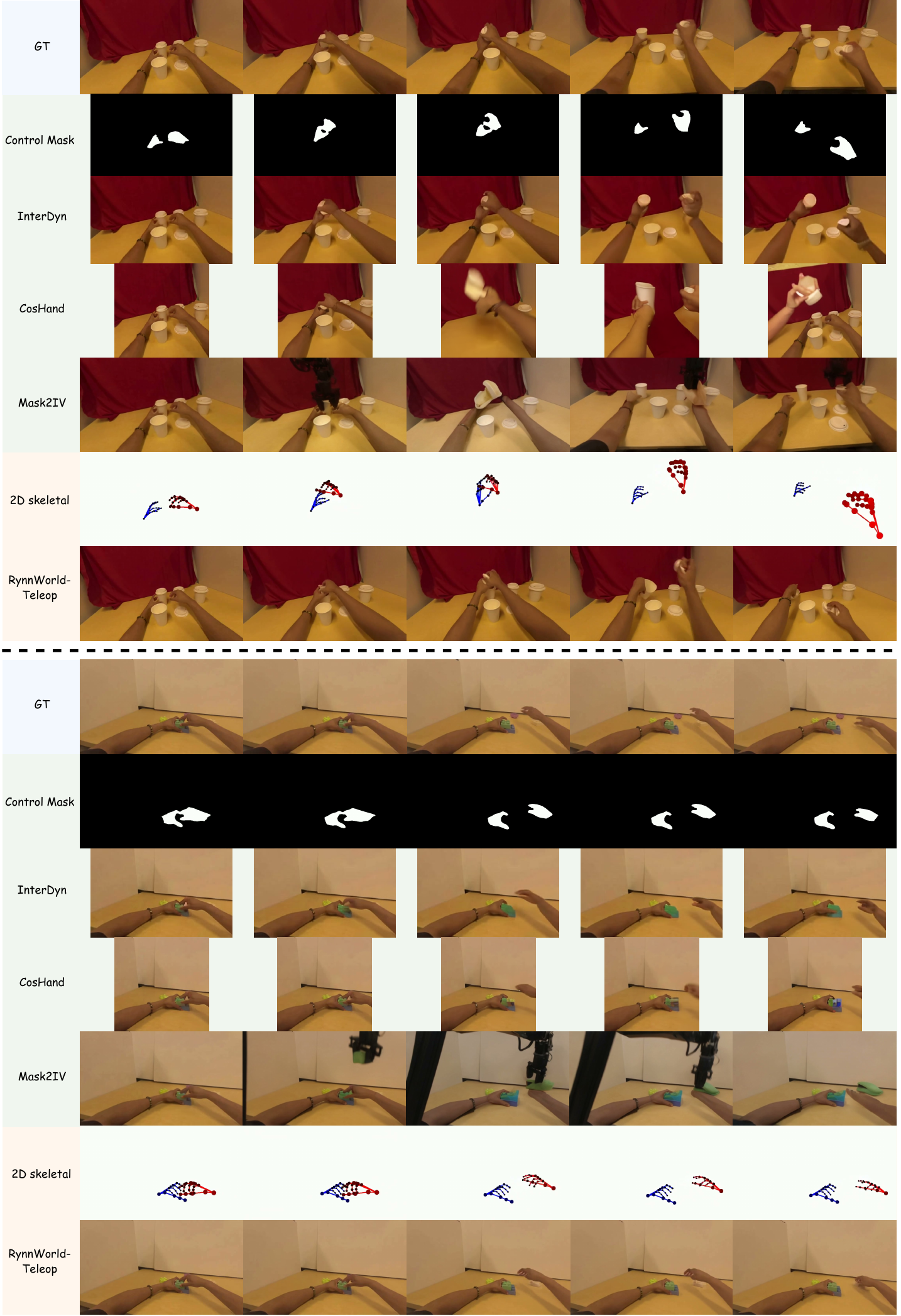}
  \caption{\textbf{Qualitative comparison.} Unlike baselines that rely on SAM-generated masks for conditioning, \our uses sparse skeletal poses, achieving superior fidelity and temporal coherence. Note that baselines are visualized at their native supported resolutions.}
  \label{fig:qualitative_comp}
\end{figure}

\begin{figure}
  \centering
    \includegraphics[width=\linewidth]{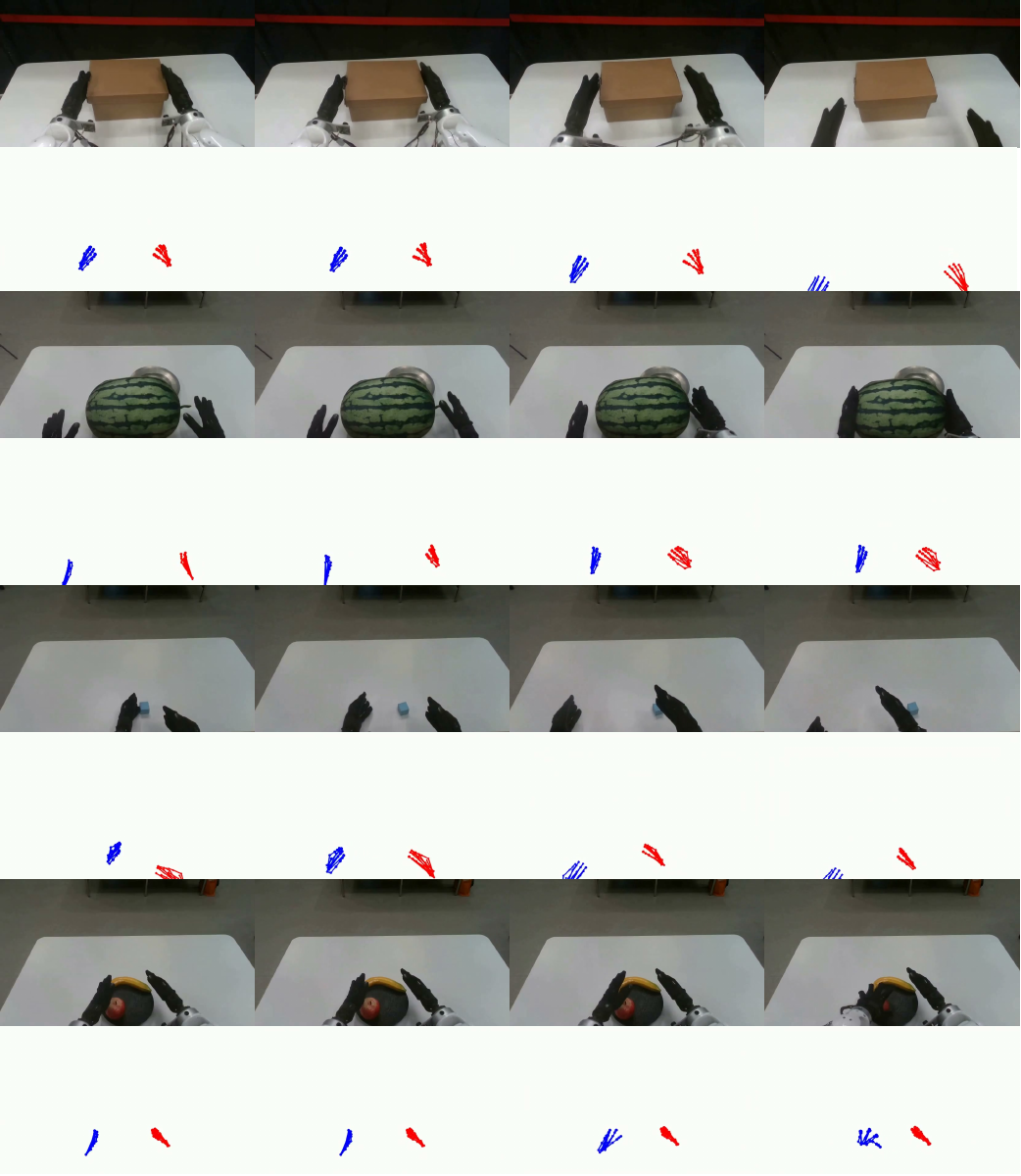}
    \caption{\textbf{Qualitative results of \our on robotic manipulation tasks.} Starting from a single reference image and a sequence of human hand-pose streams, \our synthesizes high-fidelity, temporally coherent robotic execution videos. The results demonstrate the model's ability to render complex dexterous interactions, such as bimanual coordination and high-precision object handling, while maintaining strict adherence to the input action signal.}
  \label{fig:robo-vis}
\end{figure}

\noindent
\section{Visual Comparison Results}
As shown in Fig.~\ref{fig:qualitative_comp}, \our produces more temporally stable and physically plausible interactions compared to mask-based methods. While methods like Mask2IV maintain structural shape, they often suffer from texture flickering and lack fine-grained finger articulation. In contrast, our depth-aware skeletal representation provides explicit structural constraints, allowing \our to synthesize realistic finger-object contact even in complex bimanual scenarios on the EgoDex~\citep{hoque2025egodex} benchmark.

Furthermore, we demonstrate the practical efficacy of \our in the target robotic domain in Fig.~\ref{fig:robo-vis}. By successfully transferring interaction priors from human datasets to the WUJI dexterous hand, our model synthesizes photorealistic execution videos at a high resolution. These results show that \our reproduces the intricate mechanical dynamics of the dexterous hand and the environmental reflections of the workspace across diverse tasks such as Lid Placement and Bimanual Lifting. We note that these are qualitative observations on selected sequences; we did not run a human discrimination study, so we make no claim that the generated videos are indistinguishable from real recordings.

\section{Baseline and Ablation Implementation Details}
\label{app:baselines}

\noindent
\textbf{Color-Aug (augmentation control).}
To verify that the gains in Tab.~\ref{tab:sim2real} are attributable to \our rather than to any expansion of a small training set, we construct a matched-budget photometric-augmentation control. Starting from the same 300 real episodes used by every baseline, we generate 300 additional episodes by applying per-episode photometric jitter to the RGB observation stream---brightness, contrast, saturation, and hue---while leaving the action sequence untouched. Jitter is sampled once per episode and held fixed across its frames, so that the augmented episode remains temporally coherent rather than flickering, and no geometric transform is applied, since that would break the correspondence between the observation and the retargeted action. The per-episode jitter samples a brightness, contrast, and saturation multiplier uniformly from $[0.8, 1.2]$ and a hue shift uniformly from $[-0.05, 0.05]$ of the hue circle. Each policy is then retrained on the resulting 600 episodes with the identical schedule and hyper-parameters used for the 300 Real + 300 \our condition, and evaluated under the same 35-trial protocol. This control shares the episode budget and the action distribution with our method and differs only in how the additional observations are produced.

\noindent
\textbf{w/o Depth Modulation (ablation variant).}
This variant isolates the contribution of the depth-aware rendering of Sec.~\ref{sec:method:rep}. We regenerate the entire skeletal conditioning corpus with depth modulation disabled: every joint and bone is drawn with a single fixed color and a fixed radius, so the rendered skeleton retains the same 21-joint topology, the same 2D projection, and the same line connectivity, and differs from our representation only in that camera-space depth no longer modulates appearance. The variant is then trained from the same Wan-2.2-TI2V-5B initialization with the identical two-stage LoRA schedule, learning rates, and step counts as \our (LoRA), and is evaluated on EgoDex-Test under the same protocol, so that the only difference between the two rows of Tab.~\ref{tab:exp} is the presence of the depth cue in the control signal.

\noindent
\textbf{CosHand.}
CosHand~\citep{sudhakar2024controlling} is a diffusion-based hand image generation model that synthesizes target hand appearances conditioned on a reference image and target hand masks. We use the officially released checkpoint and adopt its image-conditioned generation pipeline. For each video, we extract frame~0 as the reference image and use SAM2~\citep{ravi2025sam} to obtain per-frame binary hand segmentation masks from the ground truth video. At each timestep $t>0$, we condition on the reference image, the reference hand mask (frame~0), and the target hand mask (frame~$t$), and generate the target frame at $256\times 256$ resolution using DDIM sampling with 50 steps and a classifier-free guidance scale of 1.5. The generated frames are then resized to $480\times 832$ to match the ground truth resolution, producing 81 frames at 16\,fps per video.

\noindent
\textbf{Mask2IV.}
Mask2IV~\citep{li2026mask2iv} is a two-stage latent video diffusion model that generates videos conditioned on an input image, text prompt, and start/end hand masks. We use the officially released checkpoints for both stages. For each ground truth video, we extract SAM2~\citep{ravi2025sam} hand masks and divide the 81-frame sequence into 5 non-overlapping chunks of 16 frames. For chunk~$i$, the conditioning consists of: the ground truth frame at index $16i$ as the input image, the hand masks at frames $16i$ (start) and $16(i+1)$ (end), and a text prompt derived from the task name (\eg ``a hand pouring''). Inference is performed at $320\times 512$ resolution with 50 DDIM steps, $\eta=1.0$, classifier-free guidance scale of 1.0, and guidance rescale of 0.7. The 5 chunks are stitched sequentially to form an 80-frame output video, which is then resized to $480\times 832$ for evaluation.

\noindent
\textbf{InterDyn.}
InterDyn~\citep{akkerman2025interdyn} extends Stable Video Diffusion (SVD) with a ControlNet branch that accepts hand interaction masks as spatial conditioning. We convert each ground truth video from 16\,fps to 12\,fps via frame resampling and split it into 2 non-overlapping chunks of 28 frames.The corresponding SAM2~\citep{ravi2025sam} hand masks are similarly resampled and converted to VP9-encoded WebM format as required by InterDyn's input pipeline. For each chunk, InterDyn takes the first frame as the reference image and the 28-frame mask sequence as ControlNet conditioning, producing 14 output frames at 6\,fps (temporal sub-sampling factor of 2) with 50 denoising steps. The two chunks yield 28 prediction frames per video at a native resolution of $256\times 384$, which are upscaled to $480\times 832$ for metric computation against the ground truth.

\noindent
\textbf{H2R.}
H2R~\citep{li2025h2r} baseline refers to a policy-data augmentation setting adapted from H2R~\citep{li2025h2r}. We use H2R to synthesize the visual observation stream, while the paired robot actions are obtained using our own retargeting pipeline. H2R itself does not output the 54-DoF joint-position actions used by the policy.

Specifically, given the same source demonstrations used in our augmentation experiments, we apply the H2R pipeline to generate robotized RGB observations. The operator pose stream is then retargeted to the target robot embodiment using our pipeline, producing the same 54-DoF action labels used for \our. The resulting H2R episodes therefore share the same action labels, episode budget, temporal length, and policy-training protocol as the \our augmentation condition. This baseline is used to evaluate observation-level human-to-robot visual augmentation under our policy-learning protocol, rather than the original large-scale encoder-pretraining setting of H2R.

\section{Extended Limitations}
\label{app:limitations}
We expand here on the limitations summarized in the Conclusion.

\noindent
\textit{Observation--action consistency is validated only indirectly.} Our evidence that the synthesized observations are consistent with the paired retargeted actions is downstream and distributional: policies trained on the data work, and the generated features overlap with real ones. We have not executed the retargeted action sequences open-loop on the physical robot and compared the resulting egocentric video against the generated video frame by frame, which would measure the accumulated observation--action divergence directly. Because the actions come from an IK solver while the video comes from an independently trained generator, small residual mismatches can in principle accumulate over a rollout, and quantifying that divergence as a function of horizon is the most informative experiment we currently lack.

\noindent
\textit{The paired-data requirement bounds the scalability argument.} Stage 2 requires paired human--robot teleoperation data, and in our experiments that data covers the same tasks and environments on which we later evaluate. In its present form the paradigm therefore does not remove the need for physical hardware; it amortizes it, replacing per-episode robot time with a one-off adaptation cost that is then reused across policies, augmentation budgets, and edited scenes. Demonstrating that a world model adapted on one task set can generate useful data for a genuinely held-out task remains future work, and until that is shown the scalability claim should be read as amortization rather than elimination.

\noindent
\textit{The real-time result applies to the distilled student.} The 40\,fps figure is achieved by the causal student, which is measurably lower in fidelity than the bidirectional teacher (Robotic-Test PSNR 18.66 vs.\ 22.53; FVD 1534 vs.\ 763). The latency budget is also measured on an H100 and covers only skeletal encoding, denoising, and decoding; it excludes tracker readout, retargeting, and the robot control loop, so it is a generator throughput rather than an end-to-end human-in-the-loop latency.

\noindent
\textit{Operator yield is not yet measured.} The trajectories used in our policy experiments are produced by replaying recorded pose streams through the generator, which exercises \our as a data generation pipeline. We have not yet measured what fraction of live, human-in-the-loop digital teleoperation attempts yield usable episodes, nor characterized the failure modes (operator error, rendering artifacts, or long-horizon drift). This measurement is necessary before digital teleoperation can be claimed to match physical teleoperation in throughput.

\noindent
\textit{Physical scope.} Although depth-modulated rendering captures 3D spatial dynamics, the model occasionally struggles with complex physical phenomena such as fine-grained liquid dynamics or the manipulation of highly deformable objects. Addressing these cases will likely require richer training data covering such interactions.

\noindent
\textit{Embodiment specificity.} Bridging the embodiment gap currently requires per-platform fine-tuning, which limits the paradigm's scalability across robot fleets. We view cross-embodiment foundation world models, possibly conditioned on robot kinematic descriptors, as a promising direction.

\end{document}